\documentclass[runningheads]{llncs}
\usepackage[T1]{fontenc}
\usepackage{graphicx}
\usepackage{amsmath,amssymb}
\usepackage{hyperref}

\usepackage{cleveref}
\usepackage{adjustbox}
\usepackage{multicol}
\usepackage{multirow}
\usepackage{booktabs}
\usepackage{tikz,tikzscale}
\usepackage{xcolor,colortbl}
\usepackage{wrapfig}
\usepackage[numbers]{natbib}

\usepackage{subcaption}
\usepackage{wrapfig}
\usetikzlibrary{shapes,decorations,arrows,calc,arrows.meta,fit,positioning}
\tikzset{
    -Latex,auto,node distance =1 cm and 1 cm,semithick,
    state/.style ={ellipse, draw, minimum width = 0.7 cm},
    empty/.style ={ellipse, minimum width = 0.7 cm},
    point/.style = {circle, draw, inner sep=0.04cm,fill,node contents={}},
    bidirected/.style={dashed},
    el/.style = {inner sep=2pt, align=left, sloped},
    error/.style ={rectangle,draw, dashed, minimum width = 0.7 cm, blue},
    connected/.style = {dashed,blue}
    }

\def\mathcolor#1#{\@mathbold{#1}}
\def\@mathbold#1#2#3{%
  \protect\leavevmode
  \begingroup
    \boldmath
    \textbf{#3}%
  \endgroup
}

\newcommand{\xpm}[2]{%
  #1$\,{\scalebox{0.8}{$\pm$\,#2}}$%
}

\definecolor{blueli}{RGB}{144,213,255}
\begin{document}
\title{TabSCM: A practical Framework for Generating Realistic Tabular Data}
%
%
\author{Sven Jacob\inst{1,2,3} \and
Bardh Prenkaj \inst{2,3} \and
Weijia Shao\inst{1}\and Gjergji Kasneci \inst{2,3}}
\authorrunning{S. Jacob et al.}
%
\institute{Federal Institute for Occupational Safety and Health (BAuA), Dresden, Germany
	\email{\{jacob.sven,shao.weijia\}@baua.bund.de}\\
	 \and School of Computation, Information and Technology,
Technical University of Munich, Munich, Germany
    \and Munich Center for Machine Learning (MCML)}

%
\maketitle              
\begin{abstract}
Most tabular-data generators match marginal statistics yet ignore causal structure, leading downstream models to learn spurious or unfair patterns. We present TabSCM, a mixed-type generator that preserves those causal dependencies. Starting from a Completed Partially Directed Acyclic Graph (CPDAG) found by any causal structure discovery algorithm, TabSCM (i) orients edges to a DAG, (ii) fits root-node marginals with KDE or categorical frequencies, and (iii) learns topologically ordered structural assignments. Such assignments are achieved using conditional diffusion models for continuous variables as child nodes and gradient-boosted trees for categorical ones. Ancestral sampling yields semantically valid records and enables exact counterfactual queries. On seven public datasets, encompassing healthcare, finance, housing, environment, TabSCM matches or surpasses state-of-the-art GAN, diffusion, and LLM baselines in statistical fidelity, downstream utility, and privacy risk, while also cutting rule-violation rates and providing causally meaningful and robust conditional interventions. Because generation is decomposed into explicit equations, it runs up to 583$\times$ faster than diffusion-only models and exposes interpretable knobs for fairness auditing and policy simulation, making TabSCM a practical choice for realism, explainability, and causal soundness.
\end{abstract}

\section{Introduction}

Synthetic data is increasingly recognized as a practical solution to many of the challenges associated with real-world data: privacy constraints \cite{jordon2018pate,chen2021synthetic_Healthcare}, data sparsity \cite{esteban2017real_sparsity_1,frid2018gan_datasparsity}, access restrictions \cite{goncalves2020generation}, and fairness concerns \cite{veale2017fairer_1,xu2018fairgan_fariness2,barbierato2022methodology_fairness3}.

In regulated domains such as healthcare, finance, and education, where the availability of labeled, high-quality data is often limited, synthetic data can offer a privacy-preserving and compliance-friendly alternative to real data \cite{goncalves2020generation,jordon2018pate}.

While substantial progress has been made in generating realistic synthetic data for images, text, and time series \cite{radford2018improving,ramesh2022hierarchical,dhariwal2021diffusion,rombach2022high}, tabular data remains a uniquely challenging modality. Tabular datasets frequently encode heterogeneous data types, non-linear dependencies, and causal relationships among variables. These causal dependencies are essential for supporting counterfactual reasoning, robust decision-making, and fair model behavior in high-stakes applications such as credit risk assessment, treatment effect estimation, and resource allocation \cite{louizos2017causal,johansson2016learning}.

Despite recent advances in generative modeling for tabular data, including methods such as TabDDPM \cite{kotelnikov2023tabddpm}, TabSyn \cite{tabsyn}, TabDiff \cite{shi2024tabdiff}, and GReaT \cite{borisov2022language}, the evaluation of generated data largely focuses on statistical fidelity (e.g., marginal/conditional distributions, pairwise correlations) or utility-based metrics (e.g., downstream task performance). 
These criteria, however, are agnostic to the preservation of causal structure, which is fundamental for ensuring that synthetic data supports valid inferences and avoids reinforcing spurious patterns.

 Causal relationships enable ante-hoc interpretability by exposing the data-generating mechanisms that give rise to observed patterns, thereby supporting faithful explanations, reasoning, and transparent auditing of downstream decisions. In TabSCM, interpretability does not only stem from the inferred graph itself, but from the explicit decomposition of the joint distribution into a set of structural assignments one mechanism per variable. Each mechanism is conditioned only on its causal parents, can be inspected independently (e.g., feature attributions within a local conditional model), and can be modified or stress-tested via targeted interventions. This modular view yields actionable explanations such as which parent variables drive a change in a generated outcome.

\begin{figure}
    \centering
    \includegraphics[width=0.99\linewidth]{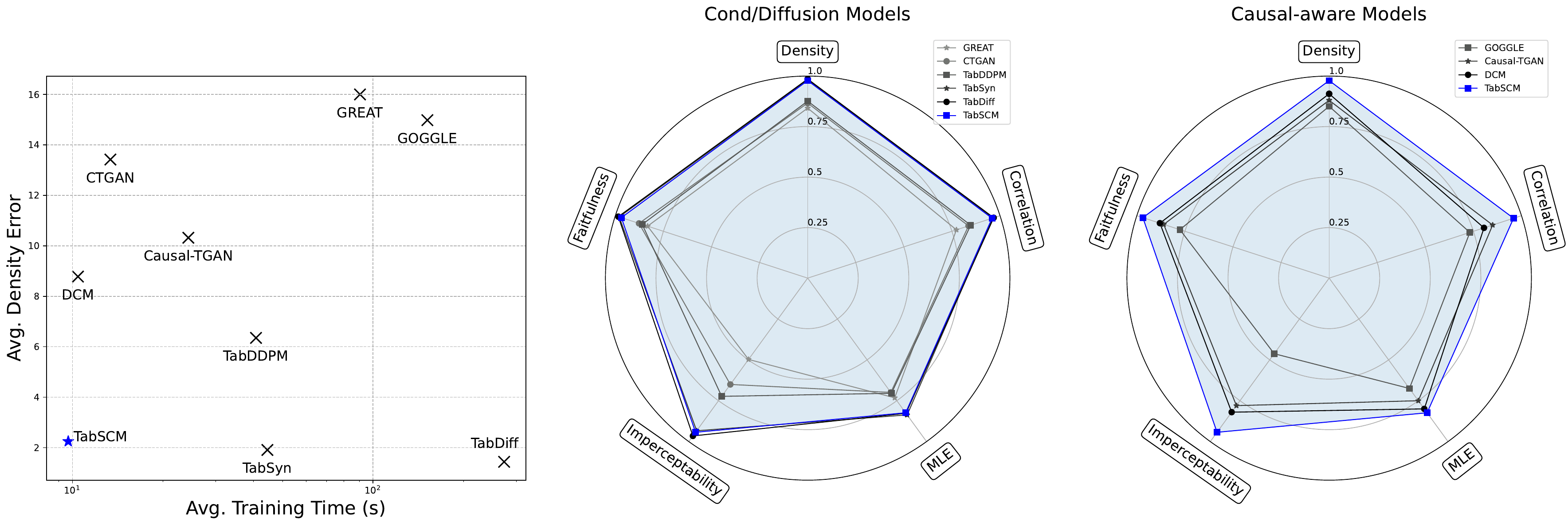}
    \caption{Shows average error (Eq.~\ref{eq:error_density}) and average training time of TabSCM against each baseline method averaged over seven real-world datasets (left).}
    \label{fig:Runtimes}
\end{figure}%

Moreover, existing state-of-the-art models often struggle to generate semantically valid samples but generate out-of-domain entries that violate known constraints, see \Cref{tab:Violations}. Such issues limit their applicability in real-world decision-making and high-stakes environments for which validity matters.

To address these limitations, we propose TabSCM, a synthetic data generation framework that leverages structural causal models (SCMs) to explicitly model and preserve the underlying causal relationships in the data. By conditioning on the causal graph, TabSCM generates samples in a topologically ordered fashion, ensuring semantic validity, supporting causal interventions, and offering greater transparency. In contrast to diffusion- or transformer-based approaches, our method is both significantly faster and more interpretable, while producing higher-quality and causally coherent synthetic data. This makes TabSCM a practical and principled tool for fair, explainable, and policy-aware data generation.

In summary, our contributions are:

(1) \textbf{A practical framework for mixed tabular data (\S\ref{sec:Method},\ref{sec:Ablation_CD},\ref{sec:Ablation_diffusion})}: We propose TabSCM, leveraging Structural Causal Models (SCMs) that combine causal reasoning with decision trees and diffusion models. By leveraging the causal structure, our method improves the computational time by up to 583$\times$ in comparison to state-of-the-art diffusion methods. TabSCMs modularity significantly decreases computational time on average while providing state-of-the-art statistical fidelity, see \Cref{fig:Runtimes}. 

(2) \textbf{Realistic, valid, and privacy-preserving synthetic data (\S\ref{sec:Experiments})}: 

Through extensive experiments, we show that our proposed method achieves competitive or better results in statistical fidelity, utility, and privacy of generated data, and excels in the validity of generated samples in contrast to diffusion-only methods.

(3) \textbf{Imbalanced learning via conditional generation (\S\ref{sec:Imb_learning})}: We show that diffusion-based generators, including TabSCM, excel in imbalanced learning settings and can serve as principled alternatives to heuristic upsampling methods. By generating minority-class samples conditioned on the causal parents, TabSCM improves minority fairness and trustworthiness while preserving the underlying data-generating mechanisms.

(4) \textbf{Mechanism-level interpretability and auditability (\S\ref{sec:XAI})}: By explicitly factorizing the generator into per-variable structural assignments, TabSCM provides mechanism-level transparency, each conditional model can be inspected, debugged, and constrained in isolation, and causal interventions offer a principled interface for producing counterfactual explanations and auditing downstream decisions.

Our evaluation shows that TabSCM is able to generate privacy-preserving and valid samples while maintaining a high utility for downstream learning tasks. Code is made available at \url{https://github.com/jsve96/TabSCM}. 

\section{Related Work}
\textbf{Causal aware generation.}
Causal aware generation integrates knowledge of cause-and-effect relationships between variables into the synthetic data generation process. Unlike traditional generative models that rely solely on statistical correlations, causal models generate data in a way that respects the underlying structural dependencies. Causal-GAN \cite{Causal_GAN17} was one of the first to incorporate causality into a Generative Adversarial Network (GAN). Causal-TGAN \cite{wen2021causal} is an adaptation tailored to tabular data. While delivering promising results, the evaluation is limited, putting the generalizability in question. GOGGLE \cite{liu2023goggle} is another framework that integrates causal graphs and generative modeling for tabular data generation. It directly learns the relational structure which is parameterized by a weighted adjacency matrix, generating realistic samples. Both methods deploy GANs. Contrarily, we use diffusion-based models to increase the sample quality in error rates, downstream utility, and privacy (see \S\ref{sec:Experiments}).

\noindent\textbf{Generative models for tabular data generation.}
Over the last years, Generative Adversarial Networks (GANs), Diffusion models, and LLMs have emerged as popular frameworks for tabular data generation. 
While GAN-based methods TableGAN \cite{park2018data}, CTGAN \cite{CTGAN}, and TVAE \cite{CTGAN} paved the way for deep learning for tabular data synthesis, they often fail to capture the full diversity of the real data distribution. With the rise of diffusion-based architectures in a wide range of deep learning applications, many diffusion-based tabular data generators such as TabSyn \cite{tabsyn}, TabDDPM \cite{kotelnikov2023tabddpm}, and TabDiff \cite{shi2024tabdiff} showed promising results in tabular data synthesis. The recent success of extracting and generating language with LLMs led to various LLM-based methods for tabular data synthesis. One early work includes GReaT \cite{borisov2022language} encoding tabular data into meaningful tokens, which are then used to finetune an LLM.

\section{Problem Setup}
Let $\mathcal{D}_{\text{real}} \in \mathbb{R}^{n \times d}$ denote a real-world tabular dataset consisting of $n$ samples and $d$ variables. Our objective is to evaluate how well synthetic data generators preserve the causal structure encoded in $\mathcal{D}_{\text{real}}$.
We represent the causal structure of the data using a causal graph:
\[\mathcal{G} = (\mathcal{V}, \mathcal{E}),\] \begin{wrapfigure}{r}{0.5\textwidth}
    \centering
    \resizebox{0.5\textwidth}{!}{\begin{tikzpicture}
  \begin{scope}[scale=0.6,local bounding box=L]
    \node[state] (x1) at (0,0) {$X_{1}$};
    \node[empty] (empty1) [right = of x1] {}; 
    \node[state] (x3) [below =of empty1] {$X_{3}$};
    
    \node[empty] (empty2) [right = of x3] {}; 
    \node[state] (x4) [right = of empty2] {$X_{4}$};
    \node[state] (x2) [above = of x4] {$X_{2}$};
    \path (x1) edge (x3);
    \path (x2) edge (x4);
    \path (x3) edge (x4);

    \node[error] (e1) [below = of x1] {$\epsilon_{1}$};
    \path (e1) edge[connected] (x1);
     \node[error] (e3) [above = of x3] {$\epsilon_{3}$};
    \path (e3) edge[connected] (x3);
    \node[error] (e2) [left = of x2] {$\epsilon_{2}$};
    \path (e2) edge[connected] (x2);
     \node[error] (e4) [below =  0.5cm of e2] {$\epsilon_{4}$};
     \path (e4) edge[connected] (x4);
  \end{scope}
  \node[right=3em of L]
    {$ \begin{cases}
        X_{1}=f_{1}(\textcolor{blue}{\epsilon_{1}})\\
        X_{2}=f_{2}(\textcolor{blue}{\epsilon_{2}})\\
        X_{3}=f_{3}(X_1,\textcolor{blue}{\epsilon_{3}})\\
        X_4 = f_{4}(X_2,X_3,\textcolor{blue}{\epsilon_{4}})
       \end{cases}$};
\end{tikzpicture}}
    \caption{Minimal example of a system of four observed variables $X_{i}$, and corresponding exogenous variables $\epsilon_i$ for $i=1,2,3,4$. The causal relationships and interactions of the observed variables are illustrated on the left-hand side (causal graph $\mathcal{G}$). On the right-hand side, we describe the SCM for the associated causal graph $\mathcal{G}$ on the left.}
    \label{fig:causal_rel_interactions}
\end{wrapfigure}
where $\mathcal{V} = \{X_1, X_2, \dots, X_d\}$ is the set of observed variables (nodes), and $\mathcal{E} \subseteq \mathcal{V} \times \mathcal{V}$ is the set of directed edges representing direct causal relationships between variables, i.e., $(X_i \rightarrow X_j) \in \mathcal{E}$ implies that $X_i$ is a direct cause of $X_j$.
For our theoretical framework, we assume the causal sufficiency condition holds (i.e., no unobserved confounders), and that the data-generating process is Markovian and faithful to a Directed Acyclic Graph (DAG) denoted by $\mathcal{G}$. 
A \textit{v-structure} is a triplet of nodes $(X_{i},X_{j},X_{k})$ such that $(X_{i} \to X_{k})$ and $(X_{j} \to X_{k})$, where the nodes $X_{i},X_{j}$ are not adjacent (i.e., not connected by and edge). A \textit{Markov equivalence class} (MEC) consists of DAGs encoding the same set of conditional independence relations among the nodes. 

A \textit{Structural Causal Model} (SCM) provides a formal framework for representing and reasoning about cause-effect relationships between variables in a system. \Cref{fig:causal_rel_interactions} illustrates a minimal example of an SCM.  Formally, an SCM is defined as a tuple
\[\mathcal{M}=(\mathcal{G},\mathcal{F},E,\mathbb{P}_{E}),\]
where $\mathcal{G}=(\mathcal{V},\mathcal{E})$ denotes a causal graph, $E=(\epsilon_{1},\dots,\epsilon_{d})$ is a set of exogenous variables (noise), $\mathcal{F}$ is a collection of structural assignments (or structural equations), where each $X_{i}$ is defined as 
\begin{equation}\label{eq:Struc_assignment}
    X_{i}:=f_{i}(\textbf{PA}_{i},\epsilon_{i}),
\end{equation}
where $\textbf{PA}_i$ denotes the set of parent variables of $X_{i}$ in $\mathcal{G}$ \cite{Pearl2000-PEACMR,peters2017elements}. The joint distribution of the exogenous variables $\mathbb{P}_{E}$ are mutually independent such that 
\[\mathbb{P}_{E}:=\prod_{i=1}^{d}\mathbb{P}(\epsilon_{i}).\]
The graph structure and the mutual independence of the exogenous variables enable the causal factorization of the joint probability distribution of the observables
\begin{equation}
    \mathbb{P}(X_1,\dots,X_d)=\prod_{i=1}^{d}\mathbb{P}(X_i|\textbf{PA}_i),
\end{equation}
factorising it into conditionals according to the structural assignments \cite{Pearl2000-PEACMR}.

\section{Proposed Method}\label{sec:Method}
Here, we describe TabSCM, a model for tabular data generation utilizing a structural causal model.
\begin{figure}[!ht]
  \begin{center}
   \includegraphics[width=0.75\linewidth]{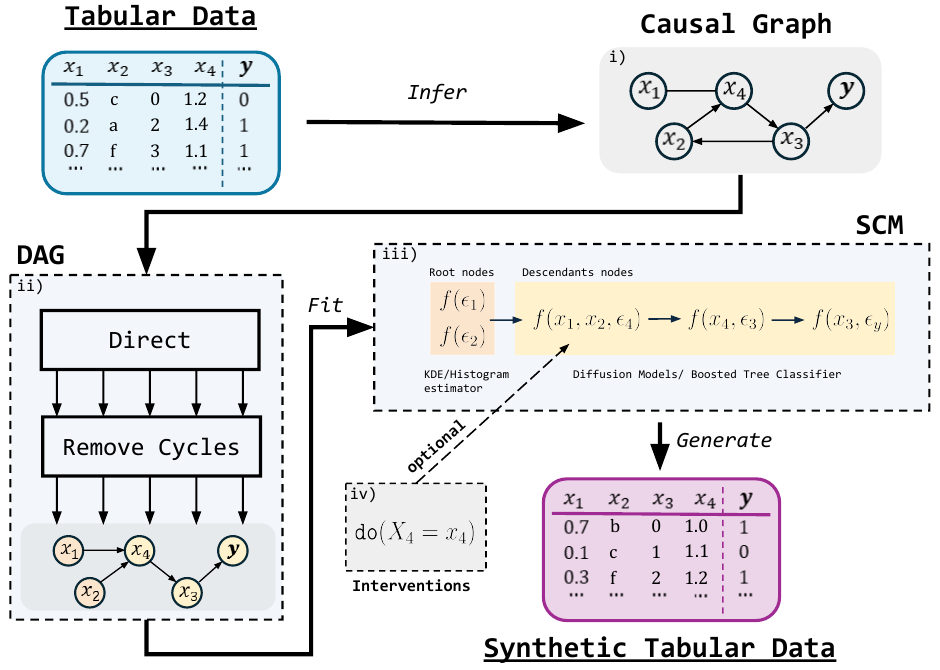}
  \end{center}
  \caption{The conceptual framework of our proposed method, including i) causal discovery, ii) refining inferred causal graph, iii) learning structural assignments and conditional sampling, and iv) Counterfactual Interventions.}
   \label{fig:conceptual_framework}
\end{figure}
Consider the topological ordering of the structural assignments (\ref{eq:Struc_assignment}) linked to each node of the underlying graph 
\[f_{\pi^{-1}(1)} \prec f_{\pi^{-1}(2)} \prec \dots \prec f_{\pi^{-1}(d)}, \]
where $\pi:\{1,\dots,d\} \to \{1,\dots,d\}$ denotes the permutation of the nodes such that $f_{\pi^{-1}(1)} \prec f_{\pi^{-1}(2)}$ indicates that node $\pi^{-1}(1)$ is a parent of node $\pi^{-1}(2)$.
Our proposed method follows the topological ordering and starts at the root nodes. We illustrate the conceptual framework of TabSCM in \Cref{fig:conceptual_framework}.

\noindent\textbf{Root nodes.} For each root node $\{X_{i}: \textbf{PA}_{i} = \emptyset; i=1,\dots,d\} \in \mathcal{V}$, we directly estimate its marginal distribution $\mathbb{P}(X_{i})$ based on the observed data $\{x_{j}^{i}\}_{j=1}^{n}$ from $ \mathcal{D}_{\text{real}}$. For continuous variables, we apply kernel density estimation (KDE) \[ \hat{\mathbb{P}}(x)=\frac{1}{nh}\sum_{j=1}^{n}K\left(\frac{x-x_j}{h}\right),\]
with a Gaussian kernel function $K(u):=\frac{1}{\sqrt{2\pi}}\exp{(\frac{-u^{2}}{2})}$. For categorical root nodes $X_{i}$, we estimate the marginal distribution with relative frequencies for $c=1,\dots,C$ with:
\[\hat{\mathbb{P}}(X_{i}=c)=\frac{1}{n}\sum_{j=1}^{n}\mathbf{1}[x_{j}^{(i)}=c].\]

\noindent\textbf{Conditional nodes.}
For each continuous non-root node $X_i \in \mathcal{V}$ with parent set $\textbf{PA}_i \neq \emptyset$, we aim to model the conditional distribution $\mathbb{P}(X_i \mid \textbf{PA}_i)$ using a conditional denoising diffusion probabilistic model (DDPM). The diffusion model defines a forward process (diffusion) that gradually adds Gaussian noise to the target variable $X_i$, and a reverse process that learns to denoise and reconstruct $X_i$ given its parents. Given a normalized input $x_i^{(0)} \sim \mathbb{P}(X_i \mid \textbf{PA}_i)$, the forward process corrupts it by gradually adding Gaussian noise $
q(x_i^{(t)} \mid x_i^{(0)}) = \mathcal{N}(x_i^{(t)}; \sqrt{\bar{\alpha}_t} x_i^{(0)}, (1 - \bar{\alpha}_t) \mathbf{I})$,
where $x_i^{(0)}$ is the clean data sample, $x_i^{(t)}$ is the noisy version at timestep $t$, and $\{\bar{\alpha}_t\}_{t=1}^T$ is the cumulative product of noise schedule coefficients \[\bar{\alpha}_t = \prod_{s=1}^t \alpha_s, \quad \alpha_s = 1 - \beta_s.\]
We train a neural network $h_\theta$ to predict the noise added at each step, conditioned on the parents $\textbf{PA}_i$ and the uniformly distributed diffusion timestep $t$,
minimizing the mean-squared error between true and predicted noise
\[\mathcal{L}_{t} = \mathbb{E}_{(x_i^{(0)}, \epsilon, t)} \left[ \left\| \epsilon - h_\theta\left(x_i^{(t)}, \textbf{PA}_i,t\right) \right\|_{2}^2 \right].\]
For categorical nodes, we deploy a boosted tree classifier \cite{friedman2001greedy} providing a flexible and robust approximation. 

\noindent\textbf{Counterfactual reasoning.} Structural Causal Model enable counterfactual reasoning, which is the ability to ask what would have happened if a variable had taken a different value. This is achieved by explicitly modeling how variables in a system interact through causal mechanisms (i.e., functional assignments that describe how each variable is generated from its causes and some noise). The key idea is that by altering these mechanisms only for specific variables, we can simulate alternative, hypothetical \textit{counterfactual} instances.
Formally, a counterfactual intervention refers to modifying the data-generating process by setting a variable $X_i \in \mathcal{V}$ to a fixed value $x_i$, and keeping the rest of the model unchanged. This is denoted by the do-operator $\text{\texttt{do}}(X_i = x_i)$, following the framework introduced by \cite{Pearl2000-PEACMR,pearl2012calculus}. The intervention replaces the structural assignment for $X_{i}$ with a fixed assignment \[X_{i}=x_{i},\]
resulting in a new SCM denoted by $\mathcal{M}_{X_{i}=x_{i}}$. Following the topological ordering, all descendant variables are generated based on the fixed assignment $X_{i}=x_{i}$ instead of the original functional assignment $f_{i}(\textbf{PA}_{i},\epsilon_{i})$. In practice, sampling from the joint distribution w.r.t. the intervention $X_{j}=x_{j}$ includes:

\noindent \textsc{\textbf{(1) Sample exogenous variables:}} Draw sample of the noise vector $E=(\epsilon_{1},\dots,\epsilon_{d}) \sim \mathbb{P}_{E}$, with i.i.d. $\epsilon_{i}$.

\noindent \textsc{\textbf{(2) Set intervention:}} Replace the structural assignment for $X_{j}$ with $X_{j}:=x_{j}$, removing the dependence on $\textbf{PA}_{j}$ and $\epsilon_{j}$. This step replaces the causal mechanism for $X_{j}$ and is the main difference to conditioning which leaves the structural assignment intact.

\noindent\textsc{\textbf{(3) Forward sampling:}} Follow the topological ordering and generate $x_{i}$ accordingly
\[x_{i}  = \begin{cases} x_{j} & \text{if }i=j\\
 x_{i} \sim \mathbb{P}(X_{i}|\textbf{PA}_{i}) & \text{otherwise} \end{cases}\]

\noindent Steps (1)--(3) generate samples from the \textbf{interventional
distribution}
\[
\mathbb{P}\bigl(\mathbf X \mid \operatorname{do}(X_j = x_j)\bigr).
\]
For \textbf{unit-level counterfactuals}, one would first condition on the factual
observation to infer the posterior
exogenous noise $\hat{E} \sim \mathbb{P}\bigl(E \mid \mathbf X = \mathbf x_{\text{obs}}\bigr)$, and then reuse the same \(\hat{E}\) in Steps (2)--(3).

\section{Experimental Setup}
\subsection{Datasets}
We use a total of seven real-world datasets covering classification and regression tasks from various application domains. The data covers large-scale datasets ($>$$250$k samples) and small-scale datasets ($<$$1000$ samples). A detailed overview is shown in \Cref{tab:my_label}

\noindent\textbf{Classification.}
The \textbf{Adult} Census Income \cite{Adult} dataset contains demographic and employment-related information from the 1994 U.S. Census to predict whether an individual earns more than \$50,000 annually. 
The Early Stage \textbf{Diabetes} Risk Prediction \cite{Early_diab} dataset contains 520 patient records collected via questionnaires at Sylhet Diabetes Hospital in Bangladesh, including 16 demographic and symptom-related features such as age, gender, polyuria, polydipsia, and sudden weight loss. The task is to predict whether an individual is diabetic.
The Home Equity Line of Credit (\textbf{HELOC}) dataset\footnote{\url{https://tinyurl.com/2r4mxjbp}}%
contains anonymized credit report features with 24 numeric variables detailing borrowers’ credit behaviors. The task is to classify whether an applicant will default (or fail to repay) their HELOC within two years.
The \textbf{Loan} dataset\footnote{\url{https://tinyurl.com/44eex2yp}}
contains over 250k historic consumer loan applications from India, including demographic, financial, and behavioral attributes. The task is to classify applicants into likely defaulters or reliable borrowers.
The \textbf{Magic} Gamma Telescope dataset \cite{magic_gamma_telescope_159} simulates the detection of high-energy gamma particles using a ground-based atmospheric Cherenkov telescope and imaging techniques. The task is to classify events as gamma-ray signals or background cosmic-ray-induced hadronic showers.

\noindent\textbf{Regression.}
The \textbf{Beijing} PM2.5 dataset \cite{Beijing} includes hourly PM2.5 measurements from the U.S. Embassy in Beijing, along with meteorological data from Beijing Capital International Airport. The goal is to predict PM2.5 concentration levels.
The California \textbf{Housing} dataset \cite{Cal_housing} contains district-level demographic and housing information across California. The task is to predict the median house value for each district.

We denote the number of numerical columns \#Num, the number of categorical columns \#Cat. \#Max Cat stands for the number of categories of the
categorical column with the most categories \cite{shi2024tabdiff}.
\begin{table*}[!ht]
    \centering
    \adjustbox{max width = 0.9\textwidth}{
    \begin{tabular}{lrrrrrrrc}
    \toprule
       Dataset  & \# Rows & \# Num & \# Cat & \# Max Cat & \# Train & \# Val & \#Test &  Taks \\
       \midrule
        Adult & $48,842$ & $6$&$9$&$42$&$28,943$&$3,618$&$16,281$& Classification\\
         Beijing & $43,824$ & $7$& $5$ & $31$& $35,058$& $4,383$& $4,383$& Regression\\
         Diabetes &$520$ &$1$&$16$&$2$&$416$&$52$&$52$& Classification\\
         HELOC &$10,459$ &$23$&$1$&$2$&$8,367$&$1,046$&$1,046$& Classification\\
         Housing &$20,640$ &$8$&$1$&$52$&$16,512$&$2,046$&$2,046$& Regression\\
         Loan & $252,000$&$2$&$10$&$317$&$201,600$&$25,200$&$25,200$& Classification\\
         Magic & $19,019$ & $10$& $1$ &$2$ &$15,215$ &$1,902$ & $1,902$ & Classification\\\bottomrule
    \end{tabular}
    }
    \vspace{0.2cm}
     \caption{Detailed overview of the datasets.}
    \label{tab:my_label}
\end{table*}
\subsection{Evaluation Metrics}
Here, we briefly describe the metrics used for each evaluation category along five dimensions i) statistical similarity, ii) downstream utility, and iii) privacy, iv) perceptibility, and v) faithfulness.
\paragraph{\textbf{Statistical Similarity:}}
We follow the two-folded evaluation of the statistical similarity proposed by SDMetrics\footnote{\url{https://docs.sdv.dev/sdmetrics}} and used by prior works \cite{tabsyn,shi2024tabdiff}. This includes column-wise density estimation and correlation error calculation. 

For each continuous column $i \in C_{\text{num}}$, we use the Kolmogorov-Smirnov (KS) test, which quantifies the maximum absolute difference between the empirical cumulative distribution functions (CDFs) of the real and synthetic data. The KS statistic is defined as
\[\text{KS}(i) := \sup_x \left| F_i^{\text{real}}(x) - F_i^{\text{syn}}(x) \right|,\]
where $F_{i}^{\text{real}},F_{i}^{\text{syn}}$ denote the CDFs of the $i$-th column of the real and synthetic data.
Lower values indicate closer alignment between the two distributions.
For categorical columns $i \in C_{\text{cat}} := C \setminus C_{\text{num}}$, we compute the Total Variation (TV) distance between empirical distributions $p_i^{\text{real}}$ and $p_i^{\text{syn}}$:
\[\text{TV}(i) := \frac{1}{2} \sum_{a \in \mathcal{A}_i} \left| p_i^{\text{real}}(a) - p_i^{\text{syn}}(a) \right|,\]
where $\mathcal{A}_i$ is the set of categories for column $i$.
Finally, following \cite{tabsyn,shi2024tabdiff}, we report the average distance 
\begin{equation}\label{eq:error_density}
    \mathbf{e}_{\text{den}}:=\frac{1}{|C|}\left(\sum_{i \in C_{\text{num}}}\text{KS(i)}+\sum_{i \in C_{\text{cat}}}\text{TV(i)}\right).
\end{equation}
To assess how well the synthetic data preserves pairwise dependencies, we compute separate errors for numerical and categorical pairs, and aggregate them into a correlation error score. This procedure is outlined in \cite{tabsyn}. For completeness, we will briefly outline the calculation in the following. We denote the set of all possible combinations of numerical columns $\mathcal{I}_\text{num}:=\{(i,j) \in C_{\text{num}}\times C_{\text{num}}\}$ for $|C_{\text{num}}| \geq 1$. 
For numerical column pairs $(i, j)$, we compute the Pearson correlation coefficients $\rho_{ij}^{\text{real}}$ and $\rho_{ij}^{\text{syn}}$, and define the numerical correlation error as
\[\mathbf{e}_{\text{corr}}^{\text{num}} := \frac{1}{|\mathcal{I}_{\text{num}}|} \sum_{(i, j) \in \mathcal{I}_{\text{num}}} \left| \rho_{ij}^{\text{real}} - \rho_{ij}^{\text{syn}} \right|.\]
For categorical column pairs $(i, j)$, we construct empirical contingency tables $R^{(ij)}$ and $S^{(ij)}$ from the real and synthetic datasets, respectively. The categorical correlation error is defined as the Total Variation distance between the normalized contingency tables:\[
\mathbf{e}_{\text{corr}}^{\text{cat}} := \frac{1}{|\mathcal{I}|-|\mathcal{I}_{\text{num}}|} \sum_{(i,j)} \frac{1}{2} \sum_{\alpha, \beta} \left| R^{(ij)}_{\alpha,\beta} - S^{(ij)}_{\alpha,\beta} \right|.\]
For mixed pairs $(i,j)$ with one numerical and one categorical column, we discretize the numerical variable into bins and apply the same procedure as for categorical pairs.
Finally, we define the overall correlation error as the average of the numerical and categorical components
\begin{equation}\label{eq:corr_error}
    \mathbf{e}_{\text{corr}} := \frac{1}{2} \left( \mathbf{e}_{\text{corr}}^{\text{num}} + \mathbf{e}_{\text{corr}}^{\text{cat}} \right).
\end{equation}
Lower values of $\mathbf{e}_{\text{corr}}$ indicate that the synthetic data preserves the correlation structure of the real data more accurately.

\paragraph{\textbf{Utility:}}
Following prior work \cite{tabsyn,shi2024tabdiff,liu2023goggle}, we assess the utility of synthetic data by training an XGBoost classifier or regressor on synthetic samples and evaluating performance on the real test set. Specifically, we split each real dataset into training and test sets, train the generative model on the real training data, and generate a synthetic dataset of equal size. An XGBoost classifier/regressor \cite{chen2016xgboost} is trained on this synthetic data using hyperparameters selected via grid search on 20 random train/validation splits, and evaluated on the real test set. We report the mean and standard deviation of AUC (for classification tasks) or RMSE (for regression tasks) across these runs. 
\paragraph{\textbf{Privacy:}}
Distance to Closest Records (DCR) is a privacy-related metric used to assess how similar synthetic data points are to real data points. 
For $x \in \mathcal{D}_{\text{Syn}}$, the DCR is \[\text{DCR}(x)=\min_{x_{j} \in \mathcal{D}_{\text{real}}}||x-x_{j}||_{1}, \]
and quantifies the distance to the nearest real record. A small value (DCR$\approx \!0$) leaks real information, a moderate value is associated with low privacy risk, and a high DCR value is safe from a privacy perspective but might have low utility in general \cite{park2018data}. Our evaluation shows that TabSCM is able to generate privacy-preserving samples while maintaining a high utility for downstream learning tasks.
\paragraph{\textbf{Perceptibility:}}
Similar to \cite{tabsyn,shi2024tabdiff}, we apply a Classifier Two Sample Test (C2ST) to quantify how difficult it is to distinguish real from synthetic data. We follow the setup given by sdmetrics\footnote{\url{https://docs.sdv.dev/sdmetrics/metrics/metrics-in-beta/detection-single-table}}, where a label gets assigned for each row of real and synthetic tabular data, both datasets are randomly split into training and validation set, a classifier is trained and evaluated on the validation set, then this procedure is repeated for different training and validation splits. The final score is based on the average $\overline{\text{AUC}}$ of the ROC across the different splits, 
\[\textsc{c2st}=1-(2\cdot\overline{\text{AUC}}-1).\]
This score is maximized $\textsc{c2st}=1$ when real and synthetic data are indistinguishable to the classifier, which corresponds to random guessing whether a sample is real or synthetic and vice versa.
\paragraph{\textbf{Faithfulness and Covering:}}
We follow prior work \cite{liu2023goggle,tabsyn,shi2024tabdiff} also evaluating data quality on a higher-order statistics assessing the models capability to capture the joint distribution. As we have previously outlined, solely evaluating MLE without context might undermine privacy, but it also ignores less informative columns. 
Therefore, we follow \cite{tabsyn} and evaluate the adopted $\alpha$-precision and $\beta$-recall \cite{alaa2022faithful}:
\begin{itemize}
    \item $\alpha$-precision: how realistic (faithful) synthetic samples are
    \item $\beta$-recall how well synthetic data covers the real data distribution.
\end{itemize}
\section{Sensitivity to Causal Discovery Algorithm}\label{sec:Ablation_CD}
The starting point of TabSCM is a Completed Partially Directed Acyclic Graph (CPDAG) modeling the describing the inter-variable relationship of the tabular data. Unveiling unknown or hidden relationships between variables is a well-known and long-studied problem in Causal Discovery. We utilize three popular approaches, PC \cite{PC}, GES \cite{GES}, and NOTEARS \cite{NOTEARS}. PC is a constraint-based method, testing conditional independencies among variables, it is lightweight and scalable, but sensitive to the choice of conditional independence test (CI-test) and threshold $\alpha$ applied. GES is a score-based method iteratively adding, removing, and reversing edges to maximize a scoring function. 

NOTEARS frames causal discovery as a continuous optimization problem, directly learning a weighted adjacency matrix representing a DAG. While PC is an efficient simple method to derive causal graphs, GES and NOTEARS can cope with complex tabular data and relationships. We use the magic dataset, which is numerical heavy and investigate how the causal discovery algorithm influences statistical fidelity and downstream utility, see \Cref{tab:CI_influence}. 

\begin{table}[!htb]
        \centering
        \adjustbox{max width = 0.99\textwidth}{
          \begin{tabular}[t]{lcccccc}
                \toprule
                &\multicolumn{3}{c}{\textbf{Density Error}} &  \multicolumn{3}{c}{\textbf{Correlation Error}}\\
                \cmidrule(l){2-4} \cmidrule(l){5-7}
               \textbf{CI-test}   & $\alpha=0.01$& $\alpha=0.05$ & $\alpha= 0.1$ & $\alpha=0.01$& $\alpha=0.05$ & $\alpha= 0.1$ \\
                \midrule
                fisherz     & \xpm{$4.44$}{$0.09$}  & \xpm{$4.12$}{$0.12$}& \xpm{$4.36$}{$0.09$}& \xpm{$7.21$}{$0.40$}  & \xpm{$7.07$}{$0.47$} & \xpm{$7.08$}{$0.80$} \\
                chiqs    &   \xpm{$4.37$}{$0.09$} & \xpm{$4.11$}{$0.19$} & \xpm{$4.26$}{$0.06$} &\xpm{$7.28$}{$0.56$}  & \xpm{$6.84$}{$0.31$} & \xpm{$7.26$}{$0.58$} \\
                \midrule
                
                 &\multicolumn{3}{c}{\textbf{AUC}} &  \multicolumn{3}{c}{\textbf{DCR}}\\
                \cmidrule(l){2-4} \cmidrule(l){5-7}
               \textbf{CI-test}   & $\alpha=0.01$& $\alpha=0.05$ & $\alpha= 0.1$ & $\alpha=0.01$& $\alpha=0.05$ & $\alpha= 0.1$ \\
                \midrule
                fisherz     & \xpm{$0.860$}{$.005$} &\xpm{$0.866$}{$.004$}&\xpm{$0.862$}{$.002$}& \xpm{$0.312$}{$.007$} &\xpm{$0.307$}{$.002$}&\xpm{$0.315$}{$.001$} \\
                chiqs    &  \xpm{$0.856$}{$.007$}&\xpm{$0.865$}{$.007$}&\xpm{$0.859$}{$.005$} &\xpm{$0.309$}{$.001$}&\xpm{$0.305$}{$.001$}&\xpm{$0.307$}{$.002$} \\
                \midrule
                \multicolumn{7}{c}{\textbf{NOTEARS}}\\
                \midrule
                & \multicolumn{3}{c}{\textbf{Density Error}} & \multicolumn{3}{c}{\textbf{Correlation Error}}\\
                \cmidrule(l){2-4} \cmidrule(l){5-7}
              & $w_{\min}=0.01$& $w_{\min}=0.2$ & $w_{\min} =  0.3$ & $w_{\min}=0.01$& $w_{\min}=0.2$ & $w_{\min} =  0.3$  \\
                \midrule
                  $\lambda_{1}=0.01$  & \xpm{$4.20$}{$0.12$}  &\xpm{$4.65$}{$0.16$} & \xpm{$4.76$}{$0.06$}   &  \xpm{$2.96$}{$0.62$}  & \xpm{$8.26$}{$0.15$}& \xpm{$8.96$}{$0.58$}\\
                $\lambda_{1} = 0.05$    & \xpm{$5.23$}{$0.06$}  & \xpm{$3.87$}{$0.08$}  & \xpm{$3.70$}{$0.05$}   & \xpm{$9.64$}{$1.12$}  & \xpm{$8.15$}{$0.49$} & \xpm{$8.25$}{$0.98$}\\
                $\lambda_{1} = 0.1$    & \xpm{$5.94$}{$0.14$}   & \xpm{$4.82$}{$0.11$} & \xpm{$6.03$}{$0.18$}  &  \xpm{$10.84$}{$0.55$}   & \xpm{$9.89$}{$0.57$} & \xpm{$11.24$}{$0.63$} \\
                \midrule
             & \multicolumn{3}{c}{\textbf{AUC}} & \multicolumn{3}{c}{\textbf{DCR}}\\
                \cmidrule(l){2-4} \cmidrule(l){5-7}
              & $w_{\min}=0.01$& $w_{\min}=0.2$ & $w_{\min} =  0.3$ & $w_{\min}=0.01$& $w_{\min}=0.2$ & $w_{\min} =  0.3$  \\
                \midrule
                $\lambda_{1} = 0.01$ & \xpm{$0.925$}{$.002$}  & \xpm{$0.903$}{$.003$} & \xpm{$0.897$}{$.005$} & \xpm{$0.227$}{$.001$}  & \xpm{$0.391$}{$.002$}& \xpm{$0.395$}{$.001$} \\
                $\lambda_{1} = 0.05$    & \xpm{$0.902$}{$.008$} & \xpm{$0.903$}{$.004$} &\xpm{$0.906$}{$.005$}  & \xpm{$0.370$}{$.001$} & \xpm{$0.391$}{$.001$} & \xpm{$0.389$}{$.001$} \\
                $\lambda_{1} = 0.1$    & \xpm{$0.906$}{$.003$} &\xpm{$0.898$}{$.004$} &\xpm{$0.903$}{$.006$} &  \xpm{$0.436$}{$.001$}& \xpm{$0.457$}{$.002$} &\xpm{$0.485$}{$.003$} \\
                \bottomrule
            \end{tabular}
            }
            \vspace{0.2cm}
            \caption{Shows the influence of threshold $\alpha$ using different conditional independence tests, and different weights $w_{\min}$ and regularizations $\lambda$ for NOTEARS on the statistical fidelity, downstream utility, and privacy.}
            \label{tab:CI_influence}
\end{table}

\section{Influence of Diffusion steps and Epochs}\label{sec:Ablation_diffusion}
In the following, we investigate how the number of epochs $n$ and the number of diffusion steps $t$ influence the statistical fidelity and downstream utility. We set $n=\{100,250,500,1000\}$ and $t=\{500,1000,1500,2000\}$ for each run. We used the same DAG, which we constructed using NOTEARS with $\lambda_{1}=0.01,w_{\min}=0.01$. We fitted TabSCM to the same training data with a unique combination of epochs and diffusion steps. After fitting TabSCM, we sampled five synthetic datasets independently. Increasing the number of epochs leads to a better density estimation error for all numbers of diffusion steps, we see the opposite behavior for the correlation estimation. Increasing the number of diffusion steps leads to a better correlation estimation for the same number of epochs.
    
\begin{figure}[!h]
    \centering
    \includegraphics[width=0.95\linewidth]{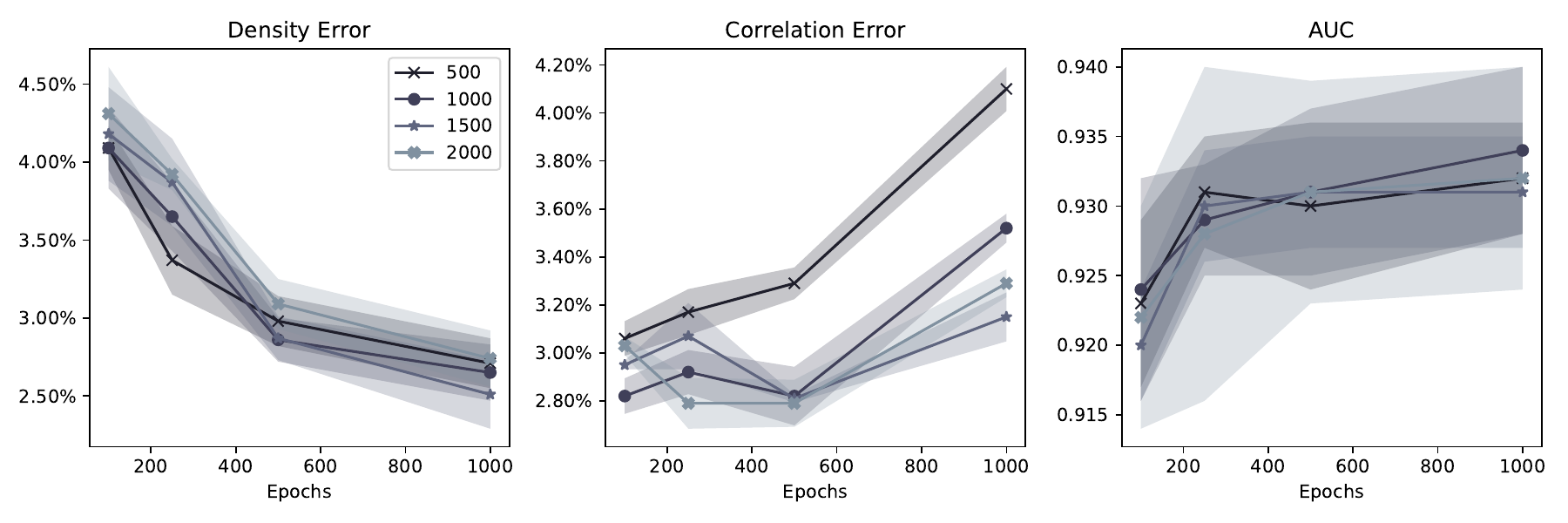}
    \caption{Shows the mean density error, correlation error, and AUC scores for different diffusion steps and for various epochs.}
    \label{fig:Ablation_1}
\end{figure}

For all parameter combinations, the AUCs are above $0.920$, such that all samples show a high utility for augmenting the training set. In contrast, the AUC score of real data is $0.948$. Increasing the number of epochs improves the AUC score in general, while the number of diffusion steps does not primarily influence AUC results for the same number of epochs, \Cref{fig:Ablation_1}. 

\begin{figure*}[!h]
    \centering
    \includegraphics[width=0.95\linewidth]{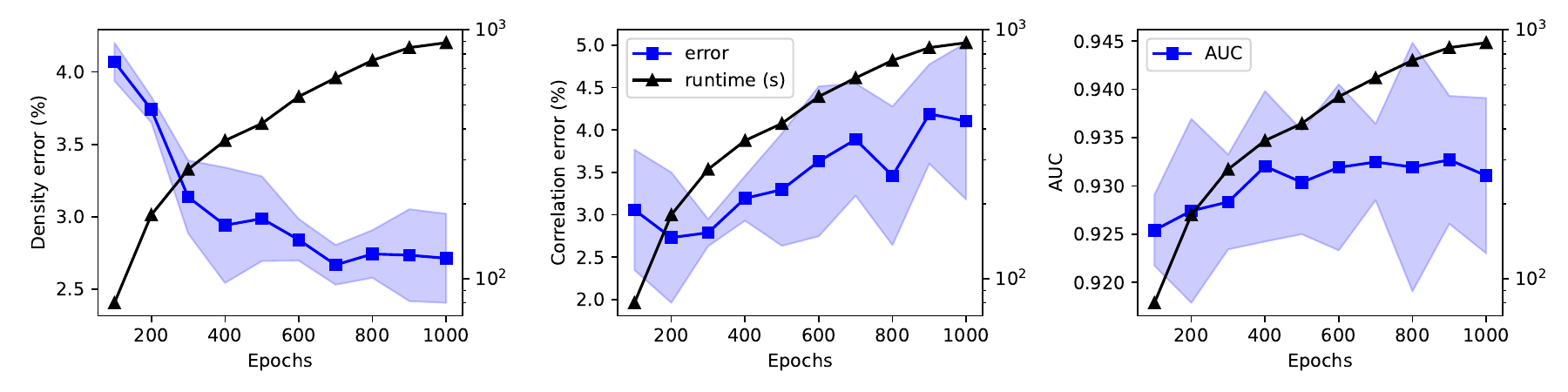}
    \caption{Illustrates how the number of epochs influences density estimation error (left), correlation error (middle), and the AUC score (right) with respect to the runtime, which is measured in seconds.}
    \label{fig:placeholder}
\end{figure*}

In the next experiment, we vary the number of epochs $n = \{100,200,\dots,1000\}$ while we fix the number of diffusion steps $t=500$. Increasing the number of epochs decreases the error of the density estimation, but this comes at a cost. This cost is runtime, which is increasing for larger epochs. We illustrate the results in \Cref{fig:placeholder}.
\section{Experimental Results}\label{sec:Experiments}
We conducted all experiments on an NVIDIA RTX 6000 Ada 48GB. For GReaT, TabDDPM, and TabSyn, we used the hyperparameters described in \cite{tabsyn}. For GOGGLE, we set the dimension of the encoder to $512$, that of the decoder to $128$, and  replaced the fixed random seed in the sampling phase. For TabDiff, we used the hyperparameters described in \cite{shi2024tabdiff}. For CTGAN, we also used the default parameter set. Whenever possible, we report the mean and standard deviation for each metric obtained from the evaluation over five trials. The choice of causal discovery method and hyperparameter setting for TabSCM is summarized in \Cref{tab:hyperparams}.
\begin{table*}[!h]
\centering
\adjustbox{max width=0.9\textwidth}{
\begin{tabular}{lrrcccc}
\toprule
\textbf{Dataset} & \textbf{Epochs} & \textbf{Diffusion steps} & \textbf{Causal Discovery} & \textbf{Weight Threshold} & \textbf{CI-Test} & \boldmath{$\alpha$} \\
\midrule
Adult     & $500$  & $500$   & GES      & --    & --        & --     \\
Beijing   & $500$  & $500$   & NOTEARS & $0.1$ & --        & --     \\
Diabetes  & $500$  & $500$   & GES      & --    & --        & --     \\
HELOC     & $1000$ & $500$   & NOTEARS & $0.1$ & --        & --     \\
Housing   & $500$  & $1000$  & GES      & --    & --        & --     \\
Loan      & $500$  & $1500$  & PC       & --    & fisherz   & $0.05$ \\
Magic     & $200$  & $2000$  & NOTEARS     & $0.01$    & --        & --     \\
\bottomrule
\end{tabular}
}
\vspace{0.2cm}
\caption{Hyperparameter settings of TabSCM for each dataset.}
\label{tab:hyperparams}
\end{table*}

\begin{table*}[!h]
\centering
\adjustbox{max width = 0.75\textwidth}{
\begin{tabular}{lrrrrrrr}
\toprule
\textbf{Method} & \textbf{Adult} & \textbf{Beijing} & \textbf{Diabetes} & \textbf{Heloc} & \textbf{Housing} & \textbf{Loan} & \textbf{Magic} \\
\midrule
GReaT             & $71.52$           & $72.70$           & $168.68$         & $96.85$          & $30.80$          & $168.93$         & $24.05$ \\
TabDDPM           & $33.25$ &$58.19$ & $35.37$         & $28.85$          & $35.75$          & $32.02$ & $30.25$ \\
TabSyn            & $36.51$           & $32.35$           & $15.99$         & $31.55$          & $25.79$ & $127.16$         & $38.72$ \\
TabDiff           & $394.27$          & $338.41$          & $115.68$        & $85.61$          & $116.22$         & $1999.51$        & $118.71$ \\
GOGGLE            & $309.09$          & $492.42$          & $1.31$ & $26.41$ & $62.89$          & OOM              & $17.63$ \\
Causal-TGAN            &  $17.42$       &    $25.05$      & $\underline{0.29}$  &$\mathcolor{blue}{7.90}$  &       $7.28$    &        $104.37$      & $7.83$\\
DCM            &     $\underline{8.51}$     &   $\underline{18.57}$        & $0.51$&$\underline{19.55}$  &  $\underline{6.32}$       &  $\underline{20.04}$          & $\underline{6.03}$ \\
\midrule
TabSCM (Ours)& $\mathcolor{blue}{7.08}$ & $\mathcolor{blue}{11.46}$ & $\mathcolor{blue}{0.20}$ & $24.15$ & $\mathcolor{blue}{5.01}$ & $\mathcolor{blue}{17.45}$ & $\mathcolor{blue}{2.44}$ \\
\bottomrule
\end{tabular}
}
\vspace{0.2cm}
\caption{Training time (in minutes) for each method across datasets. \textbf{Bold} values indicate best performance, and \underline{underlined} values are second best for each dataset.}
\label{tab:Runtime-Transposed}
\end{table*}
Following prior work \cite{kotelnikov2023tabddpm,tabsyn}, we fill missing numerical data with the average column value of the corresponding column, and introduce an additional category for missing values of categorical columns. The baseline methods transform numerical columns using a QuantileTransformer, and deploy OneHotEncoding for categorical columns. TabSCM transforms numerical data with the StandardScaler, and converts categorical columns with the LabelEncoder. For columns with many categories, using Label encoding over OneHot encoding reduces the overall data matrix.
\subsection{Statistical Fidelity \& Violations}
\textbf{TabSCM achieves SoTA performance, matching or surpassing full diffusion models while maintaining a low approximation error of the marginal distributions}. Notably, TabSCM significantly outperforms deep generative models in 3 out of 7 datasets, delivering a stable and generalizable behavior, see \Cref{tab:Stat_Table}. GOGGLE could not be applied to the large-scale Loan dataset due to memory issues, and TabDDPM was not able to generate meaningful samples for the Housing dataset. 

\begin{table*}[!htb]
\adjustbox{max width=\textwidth}{
        \centering
            \begin{tabular}{lccccccc}
                \toprule
                \textbf{Method} & \textbf{Adult} & \textbf{Beijing} & \textbf{Diabetes} & \textbf{HELOC} & \textbf{Housing} & \textbf{Loan} & \textbf{Magic} \\
                \midrule
                GReaT       & \xpm{$56.6$}{$0.03$} & \xpm{$7.21$}{$0.08$} & \xpm{$5.46$}{$0.85$} & \xpm{$12.56$}{$0.11$} & \xpm{$9.19$}{$0.09$} & \xpm{$4.05$}{$0.01$} & \xpm{$16.9$}{$0.24$} \\
                CTGAN       & \xpm{$17.4$}{$0.04$} & \xpm{$18.6$}{$0.08$} & \xpm{$7.31$}{$0.34$} & \xpm{$18.1$}{$0.07$} & \xpm{$8.79$}{$0.04$} & \xpm{$13.1$}{$0.02$} & \xpm{$10.3$}{$0.09$} \\
                TabDDPM     & \xpm{$1.10$}{$0.07$} & \underline{\xpm{$1.22$}{$0.05$}} & \xpm{$4.63$}{$0.26$} & \xpm{$2.79$}{$0.13$} & \xpm{$44.7$}{$0.05$} & \xpm{$31.1$}{$0.06$} & \xpm{$1.17$}{$0.26$} \\
                TabSyn      & \underline{\xpm{$0.84$}{$0.07$}} & \xpm{$1.29$}{$0.04$} & \xpm{$2.60$}{$0.39$} & \mathcolor{blue}{\textbf{\xpm{$2.09$}{$0.13$}}} & \mathcolor{blue}{\textbf{\xpm{$1.17$}{$0.13$}}} & \xpm{$4.59$}{$0.02$} & \underline{\xpm{$1.15$}{$0.12$}} \\
                TabDiff     & \mathcolor{blue}{\textbf{\xpm{$0.75$}{$0.02$}}} & \mathcolor{blue}{\textbf{\xpm{$1.16$}{$0.07$}}} & \underline{\xpm{$2.48$}{$0.73$}} & \underline{\xpm{$2.27$}{$0.09$}} & \underline{\xpm{$1.56$}{$0.16$}} & \underline{\xpm{$1.34$}{$0.01$}} & \mathcolor{blue}{\xpm{$0.79$}{$0.11$}} \\
                \rowcolor{lightgray!20}
                 GOGGLE      & \xpm{$13.9$}{$0.05$} & \xpm{$20.1$}{$0.08$} & \xpm{$37.1$}{$0.17$} & \xpm{$4.55$}{$0.08$} & \xpm{$10.9$}{$0.08$} & OOM & \xpm{$3.35$}{$0.09$} \\
                 \rowcolor{lightgray!20}
                Causal-TGAN & \xpm{$14.1$}{$0.28$} & \xpm{$15.6$}{$1.27$}& \xpm{$11.5$}{$0.37$}& \xpm{$19.5$}{$0.46$}&\xpm{$4.48$}{$0.16$} & \xpm{$2.73$}{$0.01$} &  \xpm{$4.36$}{$0.09$}\\
                \rowcolor{lightgray!20}
                DCM &  \xpm{$16.4$}{$0.06$}& \xpm{$3.46$}{$0.04$} & \xpm{$2.54$}{$0.52$} & \xpm{$18.2$}{$0.04$}& \xpm{$4.64$}{$0.11$} & \xpm{$1.91$}{$0.01$} & \xpm{$4.86$}{$0.10$} \\
                \midrule
                \rowcolor{lightgray!20}
                TabSCM (Ours) & \xpm{$2.46$}{$0.09$} & \xpm{$1.92$}{$0.05$} & \mathcolor{blue}{\textbf{\xpm{$1.73$}{$0.16$}}} & \mathcolor{blue}{\textbf{\xpm{$2.09$}{$0.08$}}} & \xpm{$2.36$}{$0.05$} & \mathcolor{blue}{\textbf{\xpm{$1.21$}{$0.02$}}} & \xpm{$3.90$}{$0.05$} \\
                \bottomrule
            \end{tabular}
}        
\vspace{0.2cm}
        \caption{The error rate (\%) of column-wise density estimation (Eq.~\ref{eq:error_density}). Lower values indicate a more accurate estimation (superior results). \textbf{Bold} numbers indicate best performance; \underline{underlined} values are second best results for each dataset.}
        \label{tab:Stat_Table}
\end{table*}
\begin{table*}[!htb]
\adjustbox{max width=\textwidth}{%
\centering

            \begin{tabular}{lccccccc}
                \toprule
               \textbf{Method} & \textbf{Adult} & \textbf{Beijing} & \textbf{Diabetes} & \textbf{HELOC} & \textbf{Housing} & \textbf{Loan} & \textbf{Magic} \\
                \midrule
                GReaT       & \xpm{$80.9$}{$0.09$} & \xpm{$9.82$}{$3.20$} & \xpm{$10.5$}{$0.93$} & \xpm{$8.29$}{$0.15$} & \xpm{$15.9$}{$1.73$} & \xpm{$22.06$}{$0.03$} & \xpm{$10.5$}{$0.39$} \\
                CTGAN       & \xpm{$18.3$}{$0.89$} & \xpm{$20.3$}{$0.03$} & \xpm{$13.8$}{$0.38$} & \xpm{$7.64$}{$0.07$} & \xpm{$16.4$}{$0.24$} & \xpm{$27.0$}{$0.03$} & \xpm{$11.1$}{$0.17$} \\
                TabDDPM     & \xpm{$2.11$}{$0.09$} & \xpm{$4.45$}{$0.12$} & \xpm{$20.5$}{$0.15$} & \xpm{$1.99$}{$0.21$} & \xpm{$21.5$}{$0.04$} & \xpm{$55.0$}{$0.07$} & \xpm{$1.20$}{$0.41$} \\
                TabSyn      & \underline{\xpm{$1.94$}{$0.44$}} & \underline{\xpm{$3.48$}{$0.26$}} & \underline{\xpm{$4.65$}{$0.82$}} & \mathcolor{blue}{\textbf{\xpm{$1.71$}{$0.33$}}} &  \underline{\xpm{$1.89$}{$0.21$}} & \xpm{$11.7$}{$0.03$} & \mathcolor{blue}{\xpm{$0.75$}{$0.09$}} \\
                TabDiff     & \mathcolor{blue}{\xpm{$1.59$}{$0.02$}} & \mathcolor{blue}{\xpm{$2.94$}{$0.14$}} & \mathcolor{blue}{\xpm{$3.95$}{$0.33$}} & \xpm{$1.93$}{$0.17$} & \xpm{$3.0$}{$0.03$} & \underline{\xpm{$8.83$}{$0.02$}} & \underline{\xpm{$0.81$}{$0.13$}} \\
                \rowcolor{lightgray!20}
                GOGGLE      & \xpm{$25.1$}{$0.09$} & \xpm{$46.6$}{$0.05$} & \xpm{$46.9$}{$0.24$} & \xpm{$10.8$}{$0.26$} & \xpm{$22.3$}{$0.20$} & OOM & \xpm{$9.35$}{$0.53$} \\
                \rowcolor{lightgray!20}
                Causal-TGAN & \xpm{$16.7$}{$0.78$}& \xpm{$6.62$}{$0.13$}& \xpm{$23.1$}{$1.27$}&\xpm{$21.5$}{$0.07$} & \xpm{$14.1$}{$0.95$} &\xpm{$9.53$}{$0.08$} & \xpm{$5.29$}{$0.37$} \\
                \rowcolor{lightgray!20}
                DCM & \xpm{$19.5$}{$0.09$} &\xpm{$9.37$}{$3.01$} & \xpm{$7.12$}{$0.95$} & \xpm{$45.3$}{$0.05$}& \xpm{$6.98$}{$0.97$} &\xpm{$12.5$}{$0.20$} & \xpm{$15.8$}{$0.99$} \\
                \rowcolor{lightgray!20}
                \midrule
                TabSCM (Ours) & \xpm{$5.12$}{$0.11$} & \xpm{$3.89$}{$0.05$} & \xpm{$6.75$}{$0.17$} & \underline{\xpm{$1.86$}{$0.48$}} & \mathcolor{blue}{\textbf{\xpm{$1.86$}{$0.04$}}} & \mathcolor{blue}{\textbf{\xpm{$6.62$}{$0.06$}}} & \xpm{$2.79$}{$0.53$} \\
                \bottomrule
            \end{tabular}
            }
        \vspace{0.2cm}
          \caption{The error rate (\%) (Eq.~\ref{eq:corr_error}) of correlation estimation between column distribution of real and synthetic data. \textbf{Bold} numbers indicate best performance; \underline{underlined} values are second best results for each dataset.}
          \label{tab:Stat_Table_corr}
\end{table*}

While the error of the column-wise density estimation is a valid indicator of whether the generative model is able to learn the distribution for each feature individually, the feature correlation is a key indicator of whether the generated data behaves realistically. TabSCM is able to outperform all baseline methods in 2 out of 7 datasets, notably this included the large-scale dataset loan, see \Cref{tab:Stat_Table_corr}. The derived causal graph models the relationship of each variable, therefore the quality of the correlation error of TabSCM is linked to the causal discovery algorithm. TabSCM factorizes the data-generating process into modular components, abolishing this modularity, e.g., TabSyn and TabDiff, can lead to lower correlation errors on average.

\textbf{TabSCM achieves SoTA performance, matching or surpassing full diffusion models, while enhancing privacy and providing high downstream utility}. 
Generally, a low error of the column-wise density and correlation error implies high downstream utility, see \Cref{tab:Stat_Table,tab:Stat_Table_corr,tab:MLE_Table}.
TabSCM delivers competitive downstream utility and even outperforms the diffusion-only models, e.g., TabDiff, TabSyn, and TabDDPM, which overall excel in this metric, on two datasets. While downstream utility is a valid indicator of whether generated samples may be used to replace or augment real data, it can be achieved by copying the data, which violates privacy. TabSCM delivers a strong performance considering the trade-off between utility and privacy and shows an average relative deviation of $5\%$ from the best AUC/RMSE result while increasing the DCR by $0.74$ on average.   

\begin{table*}[ht]
\centering
\begin{minipage}{0.99\textwidth}
\adjustbox{max width=\textwidth}{%
\centering
\begin{tabular}{lcccccccc}
\toprule
\multirow{2}{*}{Method} & \multicolumn{2}{c}{\textbf{Adult}} & \multicolumn{2}{c}{\textbf{Beijing}} & \multicolumn{2}{c}{\textbf{Diabetes}} & \multicolumn{2}{c}{\textbf{HELOC}} \\
\cmidrule{2-9}
& AUC $(\uparrow)$ & DCR & RMSE $(\downarrow)$ & DCR & AUC $(\uparrow)$ & DCR & AUC $(\uparrow)$ & DCR \\
\midrule
Real & $\mathbf{0.927}$ & - & $\mathbf{0.431}$ & - & $\mathbf{0.975}$ & - & $\mathbf{0.814}$ & - \\
\midrule
GReaT & \xpm{$0.768$}{$.171$} & \xpm{$7.036$}{$2.18$} & \xpm{$0.754$}{$.028$} & \xpm{$1.579$}{$0.02$} & \xpm{$0.687$}{$.113$} & \xpm{$2.26$}{$0.11$} & \xpm{$0.795$}{$.006$} & \xpm{$0.489$}{$0.01$} \\
CTGAN & \xpm{$0.885$}{$.004$} & \xpm{$1.389$}{$0.02$} & \xpm{$0.854$}{$.022$} & \xpm{$1.749$}{$0.01$} & \xpm{$0.544$}{$.009$} & \xpm{$4.94$}{$0.03$} & \xpm{$0.756$}{$.005$} & \xpm{$0.925$}{$0.02$} \\
TabDDPM & \xpm{$0.908$}{$.002$} & \xpm{$0.589$}{$0.02$} & \underline{\xpm{$0.585$}{$.005$}} & \xpm{$1.439$}{$0.01$} & \xpm{$0.368$}{$.146$} & \xpm{$5.51$}{$0.05$} & \underline{\xpm{$0.808$}{$.003$}} & \xpm{$0.603$}{$0.01$} \\
TabSyn & \underline{\xpm{$0.909$}{$.001$}} & \xpm{$0.679$}{$0.01$} & \mathcolor{blue}{\xpm{$0.568$}{$.012$}} & \xpm{$1.525$}{$0.01$} & \mathcolor{blue}{\xpm{$0.991$}{$.007$}} & \xpm{$1.13$}{$0.07$} & \xpm{$0.789$}{$.009$} & \xpm{$0.639$}{$0.01$} \\
TabDiff & \mathcolor{blue}{\xpm{$0.912$}{$.002$}} & \xpm{$0.555$}{$0.01$} & \mathcolor{blue}{\xpm{$0.568$}{$.013$}} & \xpm{$1.351$}{$0.01$} & \underline{\xpm{$0.982$}{$.011$}} & \xpm{$1.03$}{$0.06$} & \xpm{$0.798$}{$.005$} & \xpm{$0.689$}{$0.01$} \\
\rowcolor{lightgray!20}
GOGGLE & \xpm{$0.814$}{$.009$} & \xpm{$1.105$}{$0.01$} & \xpm{$1.226$}{$.013$} & \xpm{$1.855$}{$0.02$} & \xpm{$0.632$}{$.009$} & \xpm{$7.42$}{$0.02$} & \xpm{$0.378$}{$.004$} & \xpm{$0.892$}{$0.01$} \\
\rowcolor{lightgray!20}
Causal-TGAN & \xpm{$0.856$}{$.003$}& \xpm{$0.801$}{$0.01$}& \xpm{$0.747$}{$.014$}&\xpm{$1.751$}{$0.01$} &\xpm{$0.516$}{$0.03$} & \xpm{$6.24$}{$0.03$}& \xpm{$0.729$}{$.001$} &\xpm{$0.790$}{$0.01$} \\
\rowcolor{lightgray!20}
DCM & \xpm{$0.578$}{$0.03$}& \xpm{$3.13$}{$0.01$}&\xpm{$0.817$}{$0.04$} &\xpm{$1.824$}{$0.01$} &\xpm{$0.943$}{$0.04$} &\xpm{$3.44$}{$0.01$} & \xpm{$0.772$}{$.013$}& \xpm{$0.856$}{$0.01$}\\
\midrule
\rowcolor{lightgray!20}
TabSCM (Ours) & \xpm{$0.842$}{$.008$} & \xpm{$1.554$}{$0.01$} & \xpm{$0.594$}{$.014$} & \xpm{$1.729$}{$0.01$} & \xpm{$0.945$}{$.028$} & \xpm{$3.40$}{$0.04$} & \mathcolor{blue}{\xpm{$0.813$}{$.004$}} & \xpm{$0.685$}{$0.01$} \\
\bottomrule
\end{tabular}
}
\end{minipage}%
\hfill
\vspace{0.15cm}
\begin{minipage}{0.99\textwidth}
\begin{center}
\adjustbox{max width=0.8\textwidth}{%
\begin{tabular}{lcccccc}
\multirow{2}{*}{Method} & \multicolumn{2}{c}{\textbf{Housing}} & \multicolumn{2}{c}{\textbf{Loan}} & \multicolumn{2}{c}{\textbf{Magic}} \\
\cmidrule{2-7}
& RMSE $(\downarrow)$ & DCR & AUC $(\uparrow)$ & DCR & AUC $(\uparrow)$ & DCR \\
\midrule
Real & $\mathbf{0.188}$ & - & $\mathbf{0.921}$ & - & $\mathbf{0.948}$ & - \\
\midrule
GReaT & \xpm{$0.265$}{$.007$} & \xpm{$0.084$}{$0.01$} & \xpm{$0.530$}{$.002$} & \xpm{$3.136$}{$0.02$} & \xpm{$0.881$}{$.003$} & \xpm{$0.157$}{$0.01$} \\
CTGAN & \xpm{$0.352$}{$.012$} & \xpm{$0.158$}{$0.01$} & \xpm{$0.490$}{$.001$} & \xpm{$5.183$}{$0.02$} & \xpm{$0.821$}{$.008$} & \xpm{$0.362$}{$0.01$} \\
TabDDPM & \xpm{$0.594$}{$.07$} & \xpm{$2.955$}{$0.01$} & \xpm{$0.507$}{$.009$} & \xpm{$5.833$}{$0.01$} & \xpm{$0.932$}{$.002$} & \xpm{$0.199$}{$0.01$} \\
TabSyn & \mathcolor{blue}{\xpm{$0.235$}{$.004$}} & \xpm{$0.109$}{$0.01$} & \underline{\xpm{$0.561$}{$.001$}} & \xpm{$4.771$}{$0.01$} & \underline{\xpm{$0.934$}{$.005$}} & \xpm{$0.199$}{$0.01$} \\
TabDiff & \underline{\xpm{$0.241$}{$.011$}} & \xpm{$0.118$}{$0.01$} & \xpm{$0.506$}{$.001$} & \xpm{$5.423$}{$0.01$} & \mathcolor{blue}{\xpm{$0.936$}{$.006$}} & \xpm{$0.202$}{$0.01$} \\
\rowcolor{lightgray!20}
GOGGLE & \xpm{$0.381$}{$.003$} & \xpm{$0.264$}{$0.01$} & OOM & OOM & \xpm{$0.876$}{$.002$} & \xpm{$0.331$}{$0.01$} \\
 \rowcolor{lightgray!20}
Causal-TGAN & \xpm{$0.307$}{$.004$}&\xpm{$0.135$}{$0.01$} &\xpm{$0.538$}{$0.02$} & \xpm{$4.412$}{$.001$}&\xpm{$0.885$}{$.004$} & \xpm{$0.253$}{$0.01$} \\
\rowcolor{lightgray!20}
DCM & \xpm{$0.266$}{$.001$} &\xpm{$0.164$}{$0.01$} & \xpm{$0.525$}{$.012$}&\xpm{$4.356$}{$0.01$} & \xpm{$0.911$}{$.004$} & \xpm{$0.348$}{$0.01$} \\
\rowcolor{lightgray!20}
\midrule
TabSCM (Ours) & \xpm{$0.253$}{$.005$} & \xpm{$0.114$}{$0.01$} & \mathcolor{blue}{\xpm{$0.591$}{$.010$}} & \xpm{$3.556$}{$0.01$} & \xpm{$0.928$}{$.001$} & \xpm{$0.256$}{$0.01$} \\
\bottomrule
\end{tabular}
}
\end{center}
\caption{Accuracy of classifiers/regressors trained on synthetic data evaluated on real test data. \textbf{Bold} numbers indicate best performance; \underline{underlined} values are second best results for each dataset. DCR values are not highlighted.}
\label{tab:MLE_Table}

\end{minipage}

\end{table*}

\textbf{TabSCM demonstrates low violation rates outperforming full diffusion models on 3 out of 4 sanity checks}.

We conduct a sanity check to determine if the longitude and latitude pair indicate a location within the state of California. \begin{table}[!h]
\centering
\adjustbox{max width=0.55\columnwidth}{
        \begin{tabular}{lcccc}
            \toprule
            Method & (S1) & (S2) & (S3) & (S4) \\
            \midrule
            Real &$2.24$&$0.00$&$0.07$&$7.60$\\
            \midrule
            GReaT &\mathcolor{blue}{\xpm{$3.96$}{$0.12$}}&\xpm{$9.92$}{$22.2$}&\mathcolor{blue}{\xpm{$0.01$}{$0.04$}}&\xpm{$7.82$}{$0.04$}\\
            CTGAN &\xpm{$20.9$}{$0.14$}&\xpm{$55.1$}{$0.31$}&\xpm{$10.83$}{$4.24$}&\mathcolor{blue}{\xpm{$4.79$}{$0.02$}}\\
            TabDDPM &\xpm{$99.9$}{$0.05$}&\xpm{$0.38$}{$0.01$}&\xpm{$0.27$}{$0.25$}&\xpm{$28.9$}{$0.01$}\\
            TabSyn &\underline{\xpm{$6.75$}{$0.15$}}&\xpm{$0.78$}{$0.05$}&\xpm{$0.28$}{$0.25$}&\xpm{$7.43$}{$0.03$}\\
            TabDiff &\xpm{$8.62$}{$0.30$}&\underline{\xpm{$0.22$}{$0.02$}}&\xpm{$1.60$}{$1.13$}&\xpm{$7.69$}{$0.05$}\\
            \rowcolor{lightgray!20}
            GOGGLE &\xpm{$49.7$}{$0.24$}&\xpm{$68.6$}{$0.21$}&\xpm{$48.9$}{$42.59$}&OOM\\
             \rowcolor{lightgray!20}
                Causal-TGAN & \xpm{$9.95$}{$0.30$} &\xpm{$68.8$}{$0.51$} &\xpm{$2.23$}{$1.52$} & \xpm{$9.22$}{$0.06$}  \\
                \rowcolor{lightgray!20}
                DCM & \xpm{$17.2$}{$0.18$}& \xpm{$86.8$}{$0.19$} &\xpm{$50.5$}{$0.82$} & \xpm{$7.38$}{$0.05$} \\
                \rowcolor{lightgray!20}
            \midrule
            TabSCM &\xpm{$6.95$}{$0.18$}&\mathcolor{blue}{\xpm{$0.00$}{$0.00$}}&\underline{\xpm{$0.11$}{$0.10$}}&\underline{\xpm{$7.36$}{$0.05$}}\\
            \bottomrule
            \end{tabular}%
        }
        \vspace{0.2cm}
        \caption{Reports violation rates (\%) described in \Cref{sec:Experiments}. \textbf{Bold} numbers indicate best performance; \underline{underlined} values are second best results for each sanity check.}
\label{tab:Violations}
\end{table}
For the specific boundary\footnote{\url{https://github.com/PublicaMundi/MappingAPI/blob/master/data/geojson/us-states.json}} data, the real training data had an inherent violation rate of $2.24\%$. This can happen given that we collected the boundary data independently.

We conduct sanity checks of the Adult, Housing, and Loan dataset.
Additionally, we investigate if longitude and latitude imply a location inside of California (S1), education implies the same education number as given by the real data (S2), if husband or wife implies the appropriate gender (S3), and whether age implies an appropriate value of experience (S4).
We specify a mismatch between age and experience whenever experience exceeds age minus the minimum working age. Violation rates are reported in \Cref{tab:Violations}. TabSCM is able to outperform full diffusion models with lower violation rates, notably LLM-based methods, e.g., GReaT dominated this metric. The underlying transformer and attention-based architecture enhance contextualization through tokenization, enabling GReaT to generate highly realistic samples for these selected pairs of variables, closely followed by TabSCM.
\subsection{Detectability \& Faithfulness}
\textbf{TabSCM achieves SoTA performance, matching or surpassing full diffusion models, in detectability and faithfullness of synthetic samples}. Diffusion models achieve the strongest overall results in general, excelling in both fidelity ($\alpha$-precision) and diversity ($\beta$-recall) (\Cref{tab:alpha_beta_joint}) and detectability (\Cref{tab:c2st_Table}). TabSCM consistently ranks among the top methods, outperforming prior causal approaches (DCM, Causal-TGAN) and GAN-based baselines, while remaining competitive with full diffusion models across all datasets. These results highlight that explicit causal modeling enables TabSCM to strike a favorable balance between fidelity and diversity, making it a strong and reliable alternative when structural consistency and robustness are desired, even as full diffusion models remain the overall top performers in general.
\begin{table*}
        \centering
         \adjustbox{max width=0.99\textwidth}{
            \begin{tabular}{lccccccc}
                \toprule
                \textbf{Method} & \textbf{Adult} & \textbf{Beijing} & \textbf{Diabetes} & \textbf{HELOC} & \textbf{Housing} & \textbf{Loan} & \textbf{Magic} \\
                \midrule
                GReaT    &\xpm{$0.532$}{$.004$}&\xpm{$0.621$}{$.004$}&-&\xpm{$0.517$}{$.003$}&\xpm{$0.776$}{$.002$}&-&  \xpm{$.0419$}{$.004$}  \\
                CTGAN    &\xpm{$0.622$}{$.002$}&\xpm{$0.827$}{$.003$}&\xpm{$0.481$}{$.036$}&\xpm{$0.729$}{$.005$}&\xpm{$0.809$}{$.003$}&\xpm{$0.451$}{$.001$}&\xpm{$0.634$}{$.003$} \\
                TabDDPM   &\xpm{$0.955$}{$.005$}&\xpm{$0.953$}{$.003$}&\xpm{$0.949$}{$.029$}&\xpm{$0.904$}{$.009$}&\xpm{$0.04$}{$.017$}&\xpm{$0.273$}{$.002$}& \xpm{$0.989$}{$.006$}\\
                TabSyn    &\underline{\xpm{$0.979$}{$.006$}}&\xpm{$0.944$}{$.001$}&\underline{\xpm{$0.985$}{$.008$}}&\mathcolor{blue}{\xpm{$0.934$}{$.007$}}&\underline{\xpm{$0.992$}{$.005$}}&\xpm{$0.709$}{$.001$}&\underline{\xpm{$0.994$}{$.003$}} \\
                TabDiff  &\mathcolor{blue}{\xpm{$0.985$}{$.002$}}&\mathcolor{blue}{\xpm{$0.959$}{$.005$}}&\mathcolor{blue}{\xpm{$0.999$}{$.002$}}&\underline{\xpm{$0.924$}{$.007$}}&\xpm{$0.968$}{$.006$}&\mathcolor{blue}{\xpm{$0.925$}{$.002$}}&\mathcolor{blue}{\xpm{$0.998$}{$.001$}}\\ 
                \rowcolor{lightgray!20}
                GOGGLE    &\xpm{$0.112$}{$.004$}&\xpm{$0.352$}{$.002$}&\xpm{$0.001$}{$.001$}&\xpm{$0.790$}{$.008$}&\xpm{$0.712$}{$.001$}&OOM&\xpm{$0.855$}{$.005$}\\
                \rowcolor{lightgray!20}
                Causal-TGAN & \xpm{$0.706$}{$.001$}& \xpm{$0.944$}{$.010$} & \xpm{$0.398$}{$.001$} & \xpm{$0.804$}{$.004$} & \xpm{$0.912$}{$.003$}& \xpm{$0.867$}{$.002$} & \xpm{$0.937$}{$.004$} \\
                \rowcolor{lightgray!20}
                DCM &\xpm{$0.427$}{$.003$} &\xpm{$0.947$}{$.003$} & \xpm{$0.988$}{$.015$}&\xpm{$0.755$}{$.001$} & \xpm{$0.967$}{$.008$} & \xpm{$0.822$}{$.002$} &  \xpm{$0.958$}{$.003$}\\
                \rowcolor{lightgray!20}
                \midrule
                TabSCM (Ours)  &\xpm{$0.869$}{$.002$}&\underline{\xpm{$0.953$}{$.003$}}&\mathcolor{blue}{\xpm{$0.999$}{$.0001$}}&\xpm{$0.909$}{$.006$}&\mathcolor{blue}{\xpm{$0.995$}{$.004$}}&\underline{\xpm{$0.902$}{$.001$}}&\xpm{$0.889$}{$.004$}\\
                \bottomrule
            \end{tabular}
            }
            \vspace{0.2cm}
            \caption{Reports $\textsc{c2st}$ score. Higher values indicate superior imperceptibility of synthetic data. \textbf{Bold} numbers indicate best performance; \underline{underlined} values are second best results for each dataset.}
        \label{tab:c2st_Table}
\end{table*}

\begin{table*}[!h]
    \centering
    \adjustbox{max width=0.95\textwidth}{
        \begin{tabular}{lccccccc}
            \toprule
            \textbf{Method} & \textbf{Adult} & \textbf{Beijing} & \textbf{Diabetes} & \textbf{HELOC} & \textbf{Housing} & \textbf{Loan} & \textbf{Magic} \\
            \midrule
            GReaT &
            $0.56/0.49$ & $0.97/0.43$ & - &$0.89/0.46$ &$0.89/\mathbf{0.43}$ & - &$0.84/0.34$ \\
            CTGAN & $0.78/0.27$ & $0.93/0.39$ & $0.89/0.09$ & $0.96/0.11$ & $0.96/0.26$ & $0.87/0.06$ & $0.77/0.09$ \\
            TabDDPM &$0.96/\underline{0.49}$ & $\mathbf{0.99}/0.56$ &$0.87/0.06$ &$0.92/\mathbf{0.54}$ & - &$0.45/0.03$ &$\underline{0.98}/\underline{0.47}$ \\
            TabSyn &$\mathbf{0.99}/0.48$ &$\underline{0.98}/0.57$ &$\mathbf{0.98}/\underline{0.23}$ &$\mathbf{0.98}/\underline{0.49}$ &$\mathbf{0.99}/\mathbf{0.43}$ &$\underline{0.95}/0.10$ &$\mathbf{0.99}/\mathbf{0.48}$ \\
            TabDiff &$\underline{0.98}/\mathbf{0.55}$ &$\underline{0.98}/\underline{0.58}$ &$\underline{0.97}/\underline{0.23}$ &$\underline{0.97}/0.44$ &$\mathbf{0.99}/\underline{0.42}$ &$\mathbf{0.99}/0.07$ &$\mathbf{0.99}/\underline{0.47}$ \\
            \rowcolor{lightgray!20}
            GOGGLE &$0.54/0.07$ &$0.94/0.05$ &$0.37/0.03$ &$0.91/0.29$ &$0.94/0.07$ & OOM &$0.94/0.20$ \\
            \rowcolor{lightgray!20} 
            Causal-TGAN &$0.94/0.24$ &$0.98/0.48$ &$0.46/0.05$ &$0.95/0.33$ &$\underline{0.98}/0.32$ &$\underline{0.95}/\underline{0.11}$ &$0.97/0.26$ \\
            \rowcolor{lightgray!20}
            DCM &$0.71/0.05$ &$0.98/\mathbf{0.77}$ &$0.94/0.14$ &$0.81/0.11$ &$0.97/0.24$ &$0.92/\mathbf{0.12}$ &$0.97/0.24$ \\
            \midrule
            TabSCM (Ours) &$\underline{0.98}/0.28$ &$\underline{0.98}/0.50$ &$0.95/\mathbf{0.25}$ &$\mathbf{0.98}/0.45$ &$\mathbf{0.99}/0.39$ &$0.94/\mathbf{0.12}$ &$0.96/0.31$ \\
            \bottomrule
        \end{tabular}
    }
    \vspace{0.2cm}
    \caption{Comparison of $\alpha$-precision and $\beta$-recall scores. Each entry reports $(\alpha\text{-precision} / \beta\text{-recall})$. \textbf{Bold} values represents the best score, and  \underline{underlined} values are second best for each dataset. Higher values indicate superior results.}
    \label{tab:alpha_beta_joint}
\end{table*}

\subsection{Synthetic Data for Imbalanced Learning}\label{sec:Imb_learning}
We investigate the reliability of synthetic tabular data for downstream utility in the presence of class imbalance on Adult dataset (class ratio $3:1$) and HELOC dataset. For HELOC, we construct two imbalanced settings with class ratios of $5\!:\!1$ and $10\!:\!1$ (good risk performance to bad risk performance). For each setting, we treat every tabular data generator as an \emph{upsampling} method. We generate additional minority-class samples until class balance is restored.
As popular baselines, we include SMOTE \cite{chawla2002smote} and ADASYN \cite{he2008adasyn}.

To quantify whether upsampling improves decision-making for the minority class, we train a Random Forest classifier (fixed random seed) on each rebalanced training set and evaluate it on the same held-out real test set. We report the False Negative Ratio (FNR) and False Positive Ratio (FPR):
\[
\mathrm{FNR}=\frac{\mathrm{FN}}{\mathrm{FN}+\mathrm{TP}},\qquad
\mathrm{FPR}=\frac{\mathrm{FP}}{\mathrm{FP}+\mathrm{TN}}.
\]
We interpret a low FNR as \emph{minority fairness} (few missed positives for the minority class), and a low FPR as \emph{trustworthiness} (avoiding spurious positive predictions that may harm the majority class).

\begin{figure}[!h]
    \centering
    \includegraphics[width=0.85\linewidth]{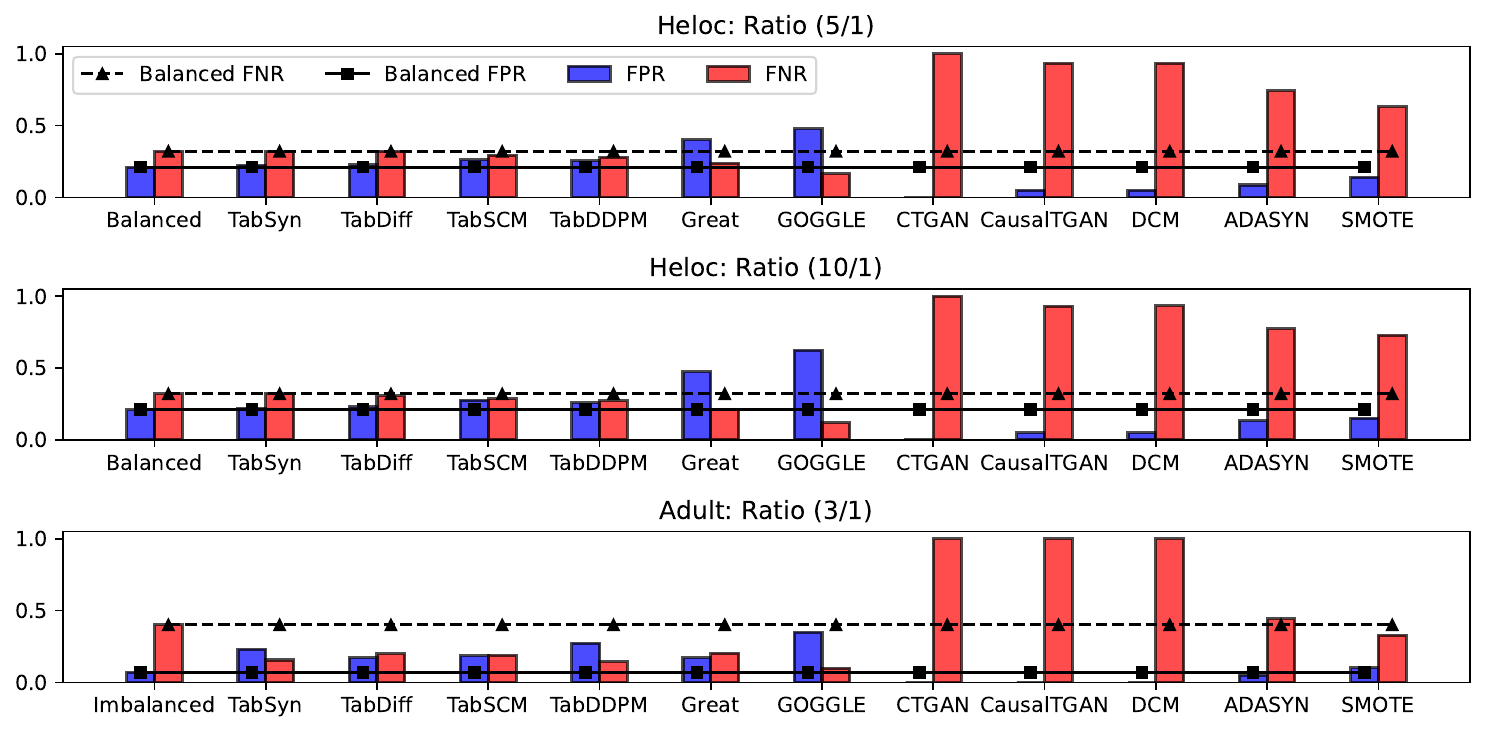}
    \caption{Shows False Negative Ratio (FNR) and False Positive Ratio (FPR) for two imbalanced scenarios of HELOC dataset using each method for synthetic upsampling and original FNR and FPR for the balanced setting.}
    \label{fig:Imbalanced_heloc}
\end{figure}

\textbf{Diffusion models including TabSCM enhance minority fairness and trustworthiness for learning under imbalanced data.}
For both class imbalance ratios of the HELOC dataset, diffusion based generator significantly outperform all other methods and recover the FNR and FPR for the (originally) balanced case. Similarly, for the class imbalanced setting of Adult, diffusion based generator have a clear edge over other baselines. Notably, TabSCM significantly outperforms other causal-aware generators including GOGGLE, Causal-TGAN, and DCM underlining its' practicability in bridging the gap between causal-aware generation and diffusion models, see \Cref{fig:Imbalanced_heloc}. 

\subsection{Mechanism-level Interpretability}\label{sec:XAI}
\begin{figure}[h]
    \centering
    \includegraphics[width=0.95\textwidth]{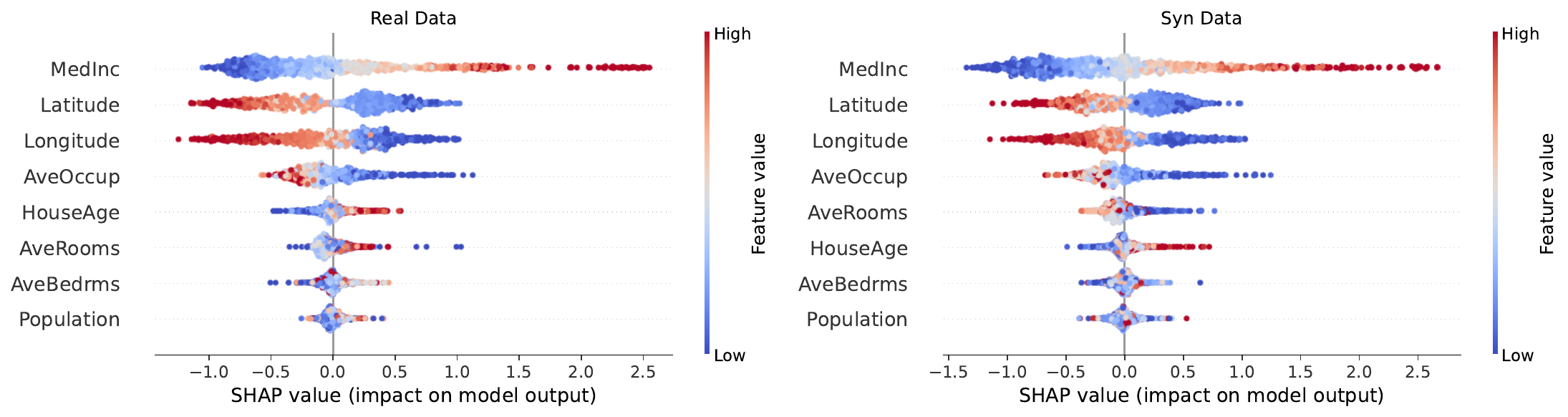}
    \caption{Shows the SHAP values of the same model trained on real data and synthetic data, generated by TabSCM.}
    \label{fig:SHAP_alignment}
    
\end{figure}

TabSCM makes generation auditable by decomposing the joint data distribution into explicit, per-variable causal mechanisms. For each variable $X_i$, it learns a separate structural assignment $X_i:=f_i(\textbf{PA}_i,\epsilon_i)$ (i.e., a conditional model $\mathbb{P}(X_i|\textbf{PA}_i)$). Because sampling follows a topological order, each mechanism can be inspected, validated, and stress-tested in isolation. We first assess whether TabSCM preserves model explanations by comparing SHAP attributions on real vs. synthetic data. Concretely, on the California Housing dataset, we train the same Gradient Boosted Regressor once on the real training set and once on a TabSCM-generated synthetic training set of equal size. We then compute SHAP values for both models on the same held-out real test instances and compare the resulting feature-attribution in \Cref{fig:SHAP_alignment}. Strong agreement indicates that TabSCM preserves the dominant predictive power of the real data.

\begin{figure}[h]
    \centering
    \includegraphics[width=0.95\textwidth]{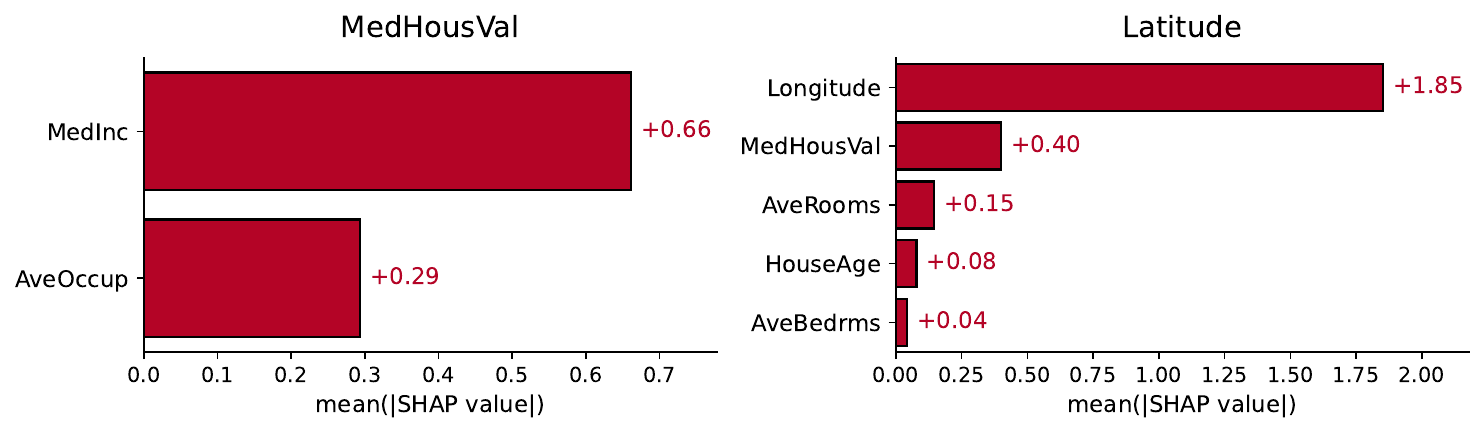}
    \caption{Shows the mean absolute SHAP values for the fitted structural assignment $f(X_c|\textbf{PA}_c)$ for $c=\text{MedHousVal}$ and  $c=\text{Latitude}$.}
    \label{fig:SHAP_TABSCM}
    
\end{figure}

Additionally, we compute mean absolute SHAP values for two representative structural assignments, $\mathbb{P}(\texttt{MedHousVal}|\textbf{PA})$ and $\mathbb{P}(\texttt{Latitude}|\textbf{PA})$, to directly inspect whether TabSCM learns plausible local mechanisms, see \Cref{fig:SHAP_TABSCM}. The resulting attributions align with domain intuition that \texttt{MedHousVal} is primarily driven by \texttt{MedInc}, which is also known to be the most influential feature for house prices in this dataset (see \Cref{fig:SHAP_alignment}. While \texttt{Latitude} is mainly explained by geographic and housing-related variables (most notably \texttt{Longitude} and \texttt{MedHousVal}).
These mechanism-level explanations provide a finer-grained and more actionable analysis than evaluating the joint distribution alone, because they reveal which parent variables actually govern each conditional generator.

As discussed earlier, TabSCM's structural causal model can be naturally used to generate counterfactual explanations. Here we show only a preliminary analysis to highlight TabSCM's versatility and leave a deeper analysis and extensions in this direction to future work.

\begin{figure}
    \centering
    \includegraphics[width=0.99\linewidth]{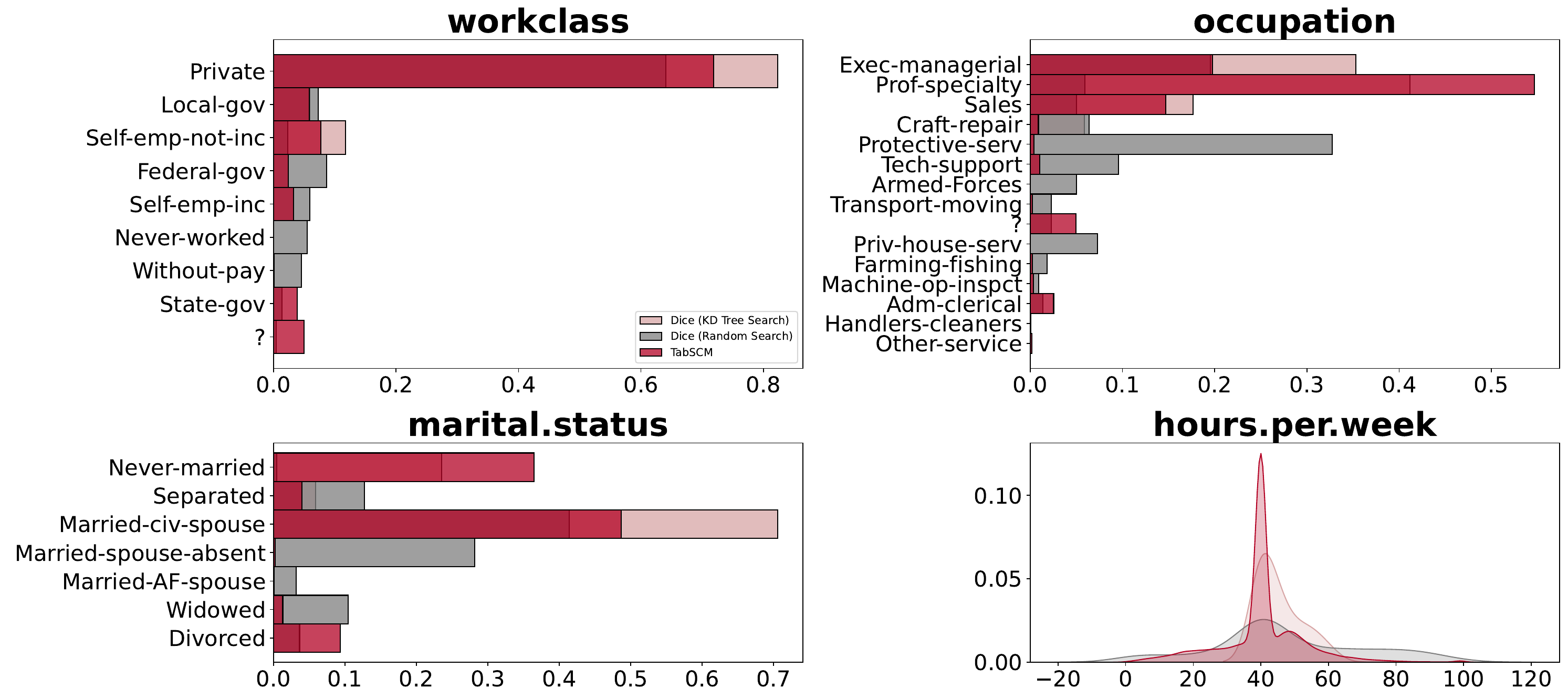}
    \caption{Shows the marginal distribution of the free attributes for counterfactual examples generated by DiCE and TabSCM.}
    \label{fig:CF_Example}
\end{figure}

We evaluate counterfactual explanations on the Adult dataset, considering \texttt{age}, \texttt{education}, \texttt{race}, \texttt{sex}, and \texttt{native-country} as protected attributes. TabSCM generates counterfactual explanations by sampling targeted interventions on the free attributes \texttt{workclass}, \texttt{marital status}, \texttt{occupation}, \texttt{relationship}, and \texttt{hours per week} and specifically intervening for the opposite class. This approach ensures that counterfactuals respect the underlying data-generating process while avoiding changes to protected attributes. We set the protected attributes to $[\texttt{35},\texttt{Masters,\texttt{White},\texttt{Male,\texttt{United-States}}}]$, and investigate which free attributes have to change to obtain an income above $\$50,000$. Therefore, we train a random forest classifier, and compare the resulting counterfactuals of TabSCM against those produced by DiCE \cite{mothilal2020explaining}, assessing the marginal distributions of the free attributes in \Cref{fig:CF_Example}. 
We selected DiCE for this preliminary analysis as it is  a widely used, model-agnostic recourse baseline that generates counterfactuals by directly optimizing feature changes to flip a classifier's prediction. 

From both methods, it becomes clear that for the given set of protected attributes, working for around $40-50$ hours as a professional expert or serving in an executive role in the private sector increases the probability of a classification of income $>\$50,000$. However, TabSCM exhibits higher certainty on this finding (see the right plot in the bottom). In addition, TabSCM, identifies ``divorce'' as a an important source for counterfactual examples (contrasting DiCE's behavior). This highlights the difference between predictor-driven counterfactuals and TabSCM’s mechanism-driven counterfactuals that follow an explicit SCM and thus better preserve plausibility under interventions and explanations.

\section{Discussion and Conclusion}

In contrast to existing generative models that prioritize marginal similarity or downstream utility, TabSCM grounds the generation process in a structural causal model (SCM) instantiated from a completed partially directed acyclic graph. By orienting this graph into a valid DAG and fitting per-variable structural assignments, using conditional diffusion models for continuous variables and gradient-boosted trees for categorical ones, TabSCM factorizes the data-generating process into modular, interpretable components.
This design ensures that samples are generated in topological order, preserving structural dependencies and avoiding rule violations by construction. Each conditional is a standalone, auditable model, which not only increases transparency but also supports fairness-aware modifications or policy-specific constraints. 
Empirical results across seven real-world datasets demonstrate that TabSCM consistently matches or outperforms state-of-the-art GAN, diffusion, and LLM-based generators in statistical fidelity, downstream predictive performance, privacy, faithfulness, and detectability. Moreover, it does so with significantly lower training time, up to $583\times$ faster than diffusion-only baselines, and with native support for semantically valid samples.
Taken together, these results show that TabSCM achieves a rare combination of realism, interpretability through causal soundness, and efficiency. It repositions SCMs from theoretical constructs to practical tools for responsible data generation, particularly in high-stakes domains such as healthcare, finance, and policy modeling. By integrating structure and causal reasoning, into generative modeling, TabSCM offers a compelling foundation for future work on fair, explainable, and intervention-aware synthetic data.

\subsubsection{\ackname} 
This research was funded by the German Federal Ministry of Labour and Social Affairs through the establishment of a Junior Research Group on Artificial Intelligence at the Federal Institute of Occupational Safety and Health (BAuA). The
presented results contribute to the development and evaluation of reliable and safe AI for industrial applications, with the overarching aim of laying the scientific foundations necessary to meet the requirements of the European Machinery Directive (2023) and the European AI Act (2024).

%
%
%
\bibliographystyle{splncs04}
\bibliography{mybib}

\end{document}